\documentclass{article}

\usepackage[main,final]{neurips_2026}

\usepackage[utf8]{inputenc}
\usepackage[T1]{fontenc}
\usepackage{textcomp}
\usepackage{microtype}
\usepackage{graphicx}
\usepackage{placeins}
\usepackage{subcaption}
\usepackage{booktabs}
\usepackage{url}
\usepackage[dvipsnames]{xcolor}
\usepackage[most]{tcolorbox}
\usepackage{amsmath}
\usepackage{amssymb}
\usepackage{amsfonts}
\usepackage{mathtools}
\usepackage{amsthm}
\usepackage{nicefrac}
\usepackage{inconsolata}
\usepackage{tabularx}
\usepackage{array}
\usepackage{algorithm}
\usepackage{algorithmic}
\usepackage{tikz}
\usetikzlibrary{arrows.meta,positioning}
\usepackage{tabularx,array,xcolor,booktabs}
\usepackage[normalem]{ulem}
\usepackage[most]{tcolorbox}
\usetikzlibrary{calc}
\definecolor{CESInk}{HTML}{263449}
\definecolor{CESMuted}{HTML}{69788C}
\definecolor{CESBorder}{HTML}{D9E1E9}
\definecolor{CESImproved}{HTML}{1F883D}
\definecolor{CESNeutral}{HTML}{9A6700}
\definecolor{CESWorsened}{HTML}{CF222E}
\definecolor{CESInsertInk}{HTML}{1A7F37}
\definecolor{CESDeleteInk}{HTML}{CF222E}

\newtcolorbox{CESExamplePanel}[3]{enhanced,
  colback=white,colframe=CESBorder,boxrule=.4pt,arc=4pt,boxsep=0pt,
  left=4pt,right=4pt,top=5pt,bottom=2pt,
  before skip=6pt,after skip=0pt,
  overlay={
    \draw[CESBorder,line width=.35pt]
      ($(interior.north west)!0.5!(interior.north east)+(0,-3pt)$) --
      ($(interior.south west)!0.5!(interior.south east)+(0,3pt)$);
    \node[anchor=west,fill=#2,text=#3,rounded corners=4pt,
      inner xsep=6pt,inner ysep=1.4pt,
      font=\sffamily\bfseries\fontsize{6.5}{7}\selectfont]
      at ([xshift=5pt]frame.north west) {#1};}}

\usepackage{hyperref}
\hypersetup{hidelinks,pdftitle={Causal Edit Serialization},pdfauthor={Louis Robinson}}
\usepackage[capitalize,noabbrev]{cleveref}
\graphicspath{{./}}
\DeclareMathOperator{\ser}{ser}
\newcommand{\CES}{\textsc{CES}}
\newcommand{\ARDSL}{\textsc{AR-DSL}}
\newcommand{\BPB}{\textsc{BPB}}
\newcommand{\CoEdit}{CoEdIT}
\newcommand{\ERRANT}{ERRANT}
\newcommand{\FIM}{FIM}

\definecolor{insgreen}{HTML}{00A000}
\definecolor{delred}{HTML}{FF0000}
\definecolor{edityellow}{HTML}{FFD000}
\definecolor{bosblue}{HTML}{2B6DFF}

\theoremstyle{plain}

\theoremstyle{definition}

\theoremstyle{remark}

\title{Causal Edit Serialization}

\author{%
  Louis Robinson \\
  Independent Researcher \\
  \texttt{louis97rf@gmail.com} \\
}

\begin{document}

\maketitle

\begin{abstract}
Standard autoregressive language models generate text sequentially, but revision requires referring back to tokens in the input context or those already generated. \emph{Causal Edit Serialization} (CES) is introduced as a framework for expressing such revisions within the same causal token stream. Special pointer tokens are paired with predicted source addresses, allowing insertion, deletion, replacement, copying and movement to be localized explicitly. A lightweight localization head and dual-index rotary position encoding support both address prediction and conditioning on previously emitted pointers, while retaining a standard causal attention mask. A pretraining procedure combines ordinary language modeling with serialized repairs derived from random corruptions, model-generated edits and repeated text. Across model scales and editing baselines, CES models demonstrate complete applied edits with modestly higher clean-text bits per byte than clean autoregressive models. Supervised and reinforcement-learning post-training provide a grammatical-error-correction proof of concept, producing more beneficial than harmful changes under reference-blind model judging. Correct repair programs also improve subsequent text prediction while the original context and its cached states are retained. CES provides a common interface for generation and localized revision, with potential extensions to iterative refinement and additional pointer-addressed transformations.
\end{abstract}

\section{Introduction}
Editing is fundamental in natural language and code: humans revise spans, replace phrases, move clauses, duplicate content, and repair earlier generations \citep{mallinson2020felixflexibletextediting,reid2022learning}. Standard autoregressive language models, however, are optimized to emit the next token from left to right \citep{radford2019language,brown2020languagemodelsfewshotlearners}. Regenerating a revised passage preserves causal decoding, but re-emits unchanged text, spending output tokens and decoding work on content already available. Emitting a localized edit avoids this repetition. The challenge is to identify the affected source spans and make those references available to subsequent predictions.

Denoising encoder--decoder models reconstruct text from corrupted inputs \citep{raffel2023exploringlimitstransferlearning,song2019massmaskedsequencesequence}. FELIX combines tagging and insertion \citep{mallinson2020felixflexibletextediting}, while insertion-based and iterative editing models relax left-to-right output construction \citep{stern2019insertiontransformerflexiblesequence,patel2025insertion,gu2019levenshteintransformer}. Here, explicit edit programs are generated within a single causal token stream. Source locations are predicted directly, and subsequent decisions condition on those references, while the standard causal attention mask is retained. Replacement, copy, and move should be available alongside insertion and deletion.

This paper presents \emph{Causal Edit Serialization} (CES). CES appends a structured edit stream after the base sequence and factors each edit into a short causal chain: predict an operation token, localize the relevant source span or destination, and then emit any payload tokens. A dual-index rotary scheme lets the model reason both about global token order and about edit addresses in the source sequence, while a lightweight localization head predicts source positions directly from the hidden states. The formulation supports insert, delete, replace, move, and copy, all inside the same decoder-only model.

The contributions are:
\begin{enumerate}
\item \textbf{A representation and training procedure for causal editing.}
CES combines source-address pointers and dual-index positional encoding with
training on random corruptions, model-generated edits, and mined copy examples.

\item \textbf{A comparison of pretrained editing models.}
Three depths and three training seeds compare CES with autoregressive edit scripts
using numeric addresses (AR-DSL, an autoregressive domain-specific language) or
relative-distance tokens, full-text regeneration, infilling with given or
predicted spans, iterative insertion/deletion, and adaptation of a clean decoder.

\item \textbf{Beneficial grammatical correction after post-training.}
Supervised fine-tuning and reinforcement learning produce positive net
improvement on grammatical error correction (GEC) under reference-blind
GPT-5 Mini judging.

\item \textbf{Continuation conditioned on edit programs.}
Supplied correct repairs improve subsequent clean-text likelihood while the
source and edit program remain in the causal history.
\end{enumerate}

\section{Related Work}
\paragraph{Insertion and iterative editing.}
The Insertion Transformer \citep{stern2019insertiontransformerflexiblesequence}, Insertion Language Models \citep{patel2025insertion}, and Levenshtein Transformer \citep{gu2019levenshteintransformer} support generation orders or refinement operations beyond appending text. FELIX \citep{mallinson2020felixflexibletextediting} combines tagging and insertion, while \citet{reid2022learning} model editing processes. EditNTS \citep{dong2019editnts} autoregressively generates edit operations with a cursor that advances through the source. CES makes source addresses explicit within the generated edit stream. The contribution of CES is the integration of explicit source addressing with causal edit generation.

\paragraph{Infilling, denoising, and patch interfaces.}
T5 \citep{raffel2023exploringlimitstransferlearning} and MASS \citep{song2019massmaskedsequencesequence} use reconstruction objectives with encoder--decoder architectures. Fill-in-the-middle training can also use a standard causal decoder by rearranging prefix, suffix, and middle \citep{bavarian2022fim}; causal attention alone therefore does not distinguish CES from FIM. The distinction is predicting source locations and heterogeneous operations together, rather than receiving a hole to fill. Learned hole selection is also tested through the predicted-span FIM baseline. LintSeq \citep{piterbarg2025lintseq} constructs synthetic edit sequences for training autoregressive models to generate code as textual diffs. Patch interfaces used in coding-agent workflows likewise separate a generated edit description from deterministic application; CES makes token addressing a learned channel inside the decoder. Agent workflows are not evaluated, and superiority over mature coding assistants is not claimed.

\paragraph{Position representations and post-training.}
Pointer networks \citep{vinyals2015pointer} use attention to select input positions. CES combines source-address prediction with conditioning on emitted pointers, partitioning rotary features \citep{su2023roformerenhancedtransformerrotary} between serialization order and source addressing. A leave-one-out policy-gradient baseline \citep{ahmadian2024rloo} is used over executed token-and-pointer trajectories in the post-training extension.

\section{Methods}
\label{sec:methods}

\subsection{Problem Setting}

Consider a sequence of standard tokens $x=(x_0,\ldots,x_{T-1})$ from a causal
language model, with prompt $x_{0:t-1}$ and generated continuation $x_{t:T-1}$.
Suppose the span $x_{i:j}$, where $t\leq i\leq j<T$, is to be replaced.
One approach is to regenerate the entire continuation with that span changed,
re-emitting the text outside the edit. An edit program can instead specify the
span and emit only its replacement. The question is how the model should
identify positions in text it has already generated.

A textual patch can locate a span by emitting short search strings around its
boundaries. Such anchors require matching rules and can be ambiguous when the
same text appears more than once. Alternatively, the model can emit an absolute
token index or a backward offset. Written as ordinary numbers, these addresses
may occupy several BPE tokens and require decoding and interpreting the number;
dedicated offset tokens instead require a predefined vocabulary of distances.
Both encode addresses through vocabulary prediction; CES instead scores
available source positions directly.

CES makes locations explicit through special pointer-role tokens paired with
predicted source addresses. The vocabulary head chooses a role, such as the
start of a replacement, and a separate localization head scores legal preceding
positions. When that token is fed back into the model, dual-index RoPE encodes
both its stream position and its source address. Thus the model both predicts
locations and conditions on them while retaining ordinary causal attention.
The source is not retroactively changed inside the decoding history: the model
appends a program, which an interpreter applies to produce the edited text.

\begin{figure}[t]
\centering
\includegraphics[width=\textwidth]{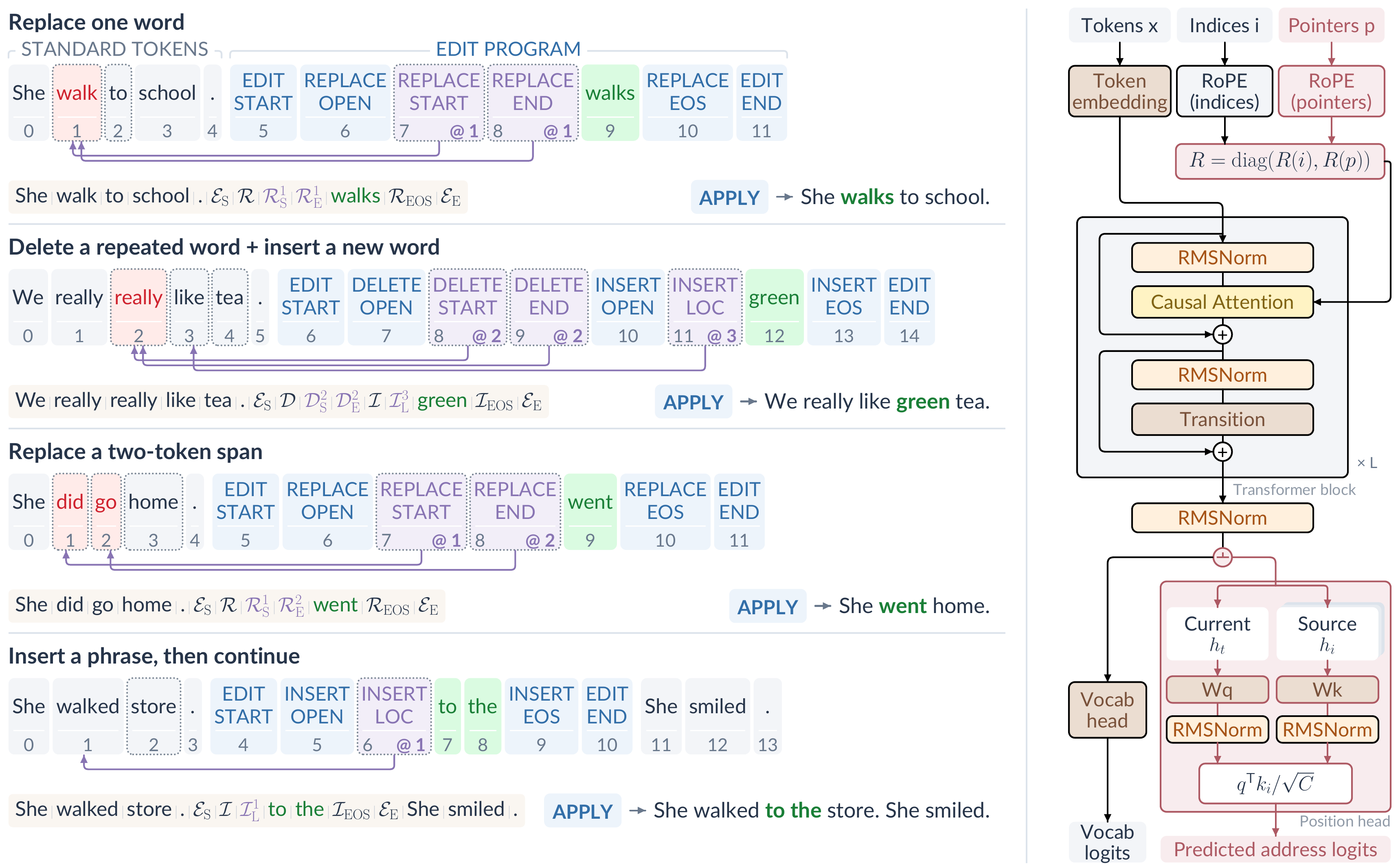}
\caption{\textbf{Causal edit programs and CES architecture.}
Left: edit programs and their applied results. Numbers denote stream positions; \texttt{@} labels give predicted source addresses. Dotted borders mark masked pretraining vocabulary targets; pointer addresses remain supervised. Right: dual-index RoPE and the localization head, with CES additions in red. $C$ denotes model width.}
\label{fig:workflow}
\end{figure}

\subsection{Edit Serialization}

Five operations are defined: $\mathcal{D}$, $\mathcal{I}$, $\mathcal{R}$,
$\mathcal{C}$, and $\mathcal{M}$, for delete, insert, replace, copy, and move.
Each edit is decomposed into smaller causal decisions: first the edit type,
then its source location or span, followed by any payload or destination
address. Each decision conditions on the preceding ones. Supplying a partial
program therefore allows the remaining decisions to be conditioned on a chosen
edit type or location.

A program containing $K$ edits is delimited by $\mathcal{E}_{\text{S}}$
and $\mathcal{E}_{\text{E}}$:
\[
\bigl(x_{0:t-1},\ \mathcal{E}_{\text{B}},\ x_{t:T-1},\
\mathcal{E}_{\text{S}},\ \ser(e_1),\ldots,
\ser(e_K),\ \mathcal{E}_{\text{E}}\bigr).
\]
The boundary marker $\mathcal{E}_{\text{B}}$ separates the protected prefix from
the editable suffix. Edits are ordered left-to-right and refer to the original
source positions, which do not shift as earlier edits are applied.

An operation symbol without a subscript denotes its opener token. The subscripts
$\text{S}$, $\text{E}$, and $\text{L}$ denote start, end, and location;
$\text{EOS}$ marks the end of a payload. Superscripts give predicted source
addresses. Figure~\ref{fig:workflow} shows the full token names alongside this
concise representation.

\begingroup
\newcommand{\CESrole}[2]{\mathcal{#1}_{\text{#2}}}
\newcommand{\CESpointer}[3]{{\color[HTML]{8874B3}\mathcal{#1}_{\text{#2}}^{#3}}}
\newcommand{\CEScomma}{\mathpunct{\textcolor{black}{,}}}
\newcommand{\CESpayload}{{\color[HTML]{1A7F37}y_1}\CEScomma\ldots\CEScomma{\color[HTML]{1A7F37}y_m}}
\[
\ser(e)=
\begin{cases}
(\mathcal{D}\CEScomma\CESpointer{D}{S}{i}\CEScomma\CESpointer{D}{E}{j}),
 & e=\operatorname{Delete}(i,j),\\[4pt]
(\mathcal{I}\CEScomma\CESpointer{I}{L}{\ell}\CEScomma\CESpayload\CEScomma\CESrole{I}{EOS}),
 & e=\operatorname{Insert}(\ell,y),\\[4pt]
(\mathcal{R}\CEScomma\CESpointer{R}{S}{i}\CEScomma\CESpointer{R}{E}{j}\CEScomma\CESpayload\CEScomma\CESrole{R}{EOS}),
 & e=\operatorname{Replace}(i,j,y),\\[4pt]
(\mathcal{C}\CEScomma\CESpointer{C}{S}{i}\CEScomma\CESpointer{C}{E}{j}\CEScomma\CESpointer{C}{L}{\ell}),
 & e=\operatorname{Copy}(i,j,\ell),\\[4pt]
(\mathcal{M}\CEScomma\CESpointer{M}{S}{i}\CEScomma\CESpointer{M}{E}{j}\CEScomma\CESpointer{M}{L}{\ell}),
 & e=\operatorname{Move}(i,j,\ell).
\end{cases}
\]
\endgroup

Purple symbols pair a token role with a predicted source address; green symbols
are standard payload tokens. Commas separate tokens in the notation.

For insertion, emitting $\mathcal{I}$ selects the insertion edit type, and
$\mathcal{I}_{\text{L}}^{\ell}$ specifies insertion after the source token
$x_\ell$ at position $\ell$.
The model then emits $y_1,\ldots,y_m$ and closes the payload with
$\mathcal{I}_{\text{EOS}}$. It can next begin another operation or finish the
edit program with $\mathcal{E}_{\text{E}}$.

For deletion, $\mathcal{D}$ is followed by
$\mathcal{D}_{\text{S}}^{i}$ and $\mathcal{D}_{\text{E}}^{j}$, identifying the
first and last tokens to remove. The span $[i,j]$ is inclusive and no payload is
needed. Replacement uses the same span description followed by its new content.

Copy retains its source span; move removes it and reinserts it elsewhere.
Their destination-position tokens, $\mathcal{C}_{\text{L}}$ and $\mathcal{M}_{\text{L}}$, specify where the span is inserted. The boundary marker also allows
insertion before the first editable token. Grammar masks restrict illegal
addresses and overlapping spans.

The illustrated programs use $\mathcal{I}_{\text{EOS}}$ and
$\mathcal{R}_{\text{EOS}}$ to mark payload completion. Including these
terminators is a design choice: they separate payload completion from the
choice of the next operation, whereas a convention without them uses the next
operation opener or edit-end token to delimit the payload.

\subsection{Dual-Index Positional Encoding}

Each token at stream position $i$ carries a global index $i$ and an address
index $p_i$. Pointer-role tokens use their source address; insertion and
replacement payloads inherit the insertion location and replacement start,
respectively. Other tokens use their global index. RoPE
\citep{su2023roformerenhancedtransformerrotary} is applied to both indices using two disjoint
channel groups, one for generation order and one for source-address relations.

For each attention head of width $d_h$, its query and key are split into equal
halves, $q_i=[q_i^{\mathrm{seq}};q_i^{\mathrm{addr}}]$ and
$k_i=[k_i^{\mathrm{seq}};k_i^{\mathrm{addr}}]$. The two RoPE matrices act on
these halves independently:
\[
R_i=\operatorname{diag}\!\bigl(R(i),R(p_i)\bigr),\qquad
\widetilde q_i=R_iq_i,\qquad \widetilde k_i=R_ik_i,
\]
where each block is a $d_h/2$-dimensional rotation. After Q/K RMS normalization
and scaling, these rotated queries and keys enter ordinary causal attention
\citep{vaswani2017attention};
values are unchanged. Their dot product combines generation-order and
source-address relations. Concatenation restores the original head width
$d_h$. The pretrained-decoder adaptation baseline
instead appends an address branch (Appendix~\ref{sec:baselinespecs}).

\subsection{Edit Localization Head}

In addition to the vocabulary head, a pointer head scores legal source positions. With projected queries $q_t=\operatorname{RMSNorm}(W_qh_t)$ and keys $k_i=\operatorname{RMSNorm}(W_kh_i)$, normalized across channels without learned affine parameters, it computes
\begin{equation}
 p(i\mid h_t,\mathcal S_t)=
 \frac{\exp(q_t^\top k_i/\sqrt{d})}{\sum_{j\in\mathcal S_t}\exp(q_t^\top k_j/\sqrt{d})},
 \quad i\in\mathcal S_t,
\end{equation}
where $\mathcal S_t$ contains legal preceding source addresses for the current grammar state. The vocabulary head predicts the pointer's role token; the address channel predicts its coordinate. This requires two small learned projections and an auxiliary source-key cache. The head and its dual-index encoding add 0.34--0.41\% parameters relative to the regeneration models at the tested depths.

\subsection{Training Data}

Edit-training streams are constructed from ClimbMix text
\citep{diao2025climb}. A modified source is paired with a causal program that
recovers its target. Random corruptions insert, delete, replace, or move token
spans. Their spacing and lengths are sampled from geometric distributions
under a token budget; inserted content mixes spans from other documents or
the current sequence with randomly sampled tokens.

To add more naturalistic revisions, localized edits proposed
by GPT-5 Mini are also serialized. A separate copy-mining pipeline removes later occurrences of
repeated spans and trains the model to restore them by referring to the earlier
occurrence. Complementary edit types and lengths are provided by these sources; the
training settings are selected through source-mixture and corruption sweeps.
Appendix~\ref{sec:corruption} gives the sampling procedure, corpus provenance,
operation counts, and quality checks.

\subsection{Training Objective}
\label{sec:trainingobjective}

The model is trained with a standard autoregressive cross-entropy loss over the serialized token stream. For location-bearing tokens, an additional cross-entropy loss is applied over the address distribution produced by the localization head. Writing each component as a mean over its supervised targets, the objective is
\[
\mathcal{L} = \frac{\sum_t \delta_t\mathcal{L}_{\text{token}}(t)}{\sum_t\delta_t} + \lambda_{\mathrm{addr}}\frac{\sum_{t \in \mathcal{A}} \mathcal{L}_{\text{addr}}(t)}{|\mathcal A|},
\]
where $\mathcal{A}$ denotes the set of address-predicting tokens and $\delta_t$ masks out corrupted source tokens whose next-token targets no longer lie on the clean language-modeling distribution. The address term is zero for a batch without pointer targets. $\lambda_{\mathrm{addr}}=0.01$ is selected from the three-seed sweep. In practice the address-loss weight is important: too much pressure on the position head hurts base language modeling, while too little pressure weakens edit localization.

\paragraph{Applying edits and continuing generation.}
\label{sec:inference}
The interpreter applies a completed program to the original source coordinates
and returns the edited text. Subsequent generation can retain the
source-plus-program history and its existing KV cache, or use the materialized
edited text as its prefix. The latter requires recomputing causal states from
the first change; the former preserves the cache but requires the model to
interpret the appended edits (Section~\ref{sec:continuationmain}).
Appendix~\ref{sec:inference_details} details edit decoding and cache management; Appendix~\ref{sec:systemscomparison} reports measured editing and continuation costs.

\subsection{GEC Post-Training}
\label{sec:gecposttraining}

The CES model is further trained for grammatical error correction without changes to its
architecture or edit language. Balanced supervised fine-tuning mixes edit
programs, clean no-op examples, and ordinary continuations. Broad
corruption-recovery RL then precedes GEC-focused RL on localized corruptions of
clean CoEdIT sentences. Sampled programs are executed, and recovery of the
known target is rewarded using edit-distance reduction and exact recovery, with penalties
for invalid or excessive edits. Policy updates use leave-one-out REINFORCE
\citep{ahmadian2024rloo} over vocabulary and pointer decisions, alongside
supervised losses. Appendix~\ref{sec:rlrecipe} details the training and evaluation
procedures; Section~\ref{sec:rl} reports the resulting editing performance.

\section{Results}
\label{sec:results}

\subsection{Repair Task and Baselines}
\label{sec:repairtask}
For controlled repair, a clean passage $y^*$ from the held-out ClimbMix validation shard is synthetically corrupted to give source $x$. Each model receives $x$ in its native prompt format and produces an edit program or revised text. Programs are applied to $x$, regenerated text is used directly, and FIM completions are inserted into their selected spans. The resulting $\hat y$ is compared with $y^*$. The target is withheld during decoding, except that oracle-span FIM receives target-derived span boundaries.

Table~\ref{tab:baseline_designs} summarizes eight NanoChat-based editing methods \citep{nanochat}. Three depths (12/18/24) and three seeds give 72 model runs, evaluated before and after supervised CoEdIT fine-tuning. CES has approximately 287M/705M/1.389B parameters. Clean AR provides additional language-model references. Exact baseline specifications are in Appendix~\ref{sec:baselinespecs}; budgets and evaluation protocols are in Appendix~\ref{sec:reproducibility}.

\subsection{Experimental Settings}
The main comparison uses mixed random and GPT-generated edits, corruption
control $1.0$, address-loss weight $\lambda_{\mathrm{addr}}=0.01$, and greedy
decoding. These settings balance language modeling and repair;
Appendix~\ref{sec:supervision} reports the source-mixture, address-loss-weight, and
corruption-exposure ablations;
Appendix~\ref{sec:samplingpolicy} compares decoding policies.

\begin{figure}[!t]
\centering
\includegraphics[width=\textwidth]{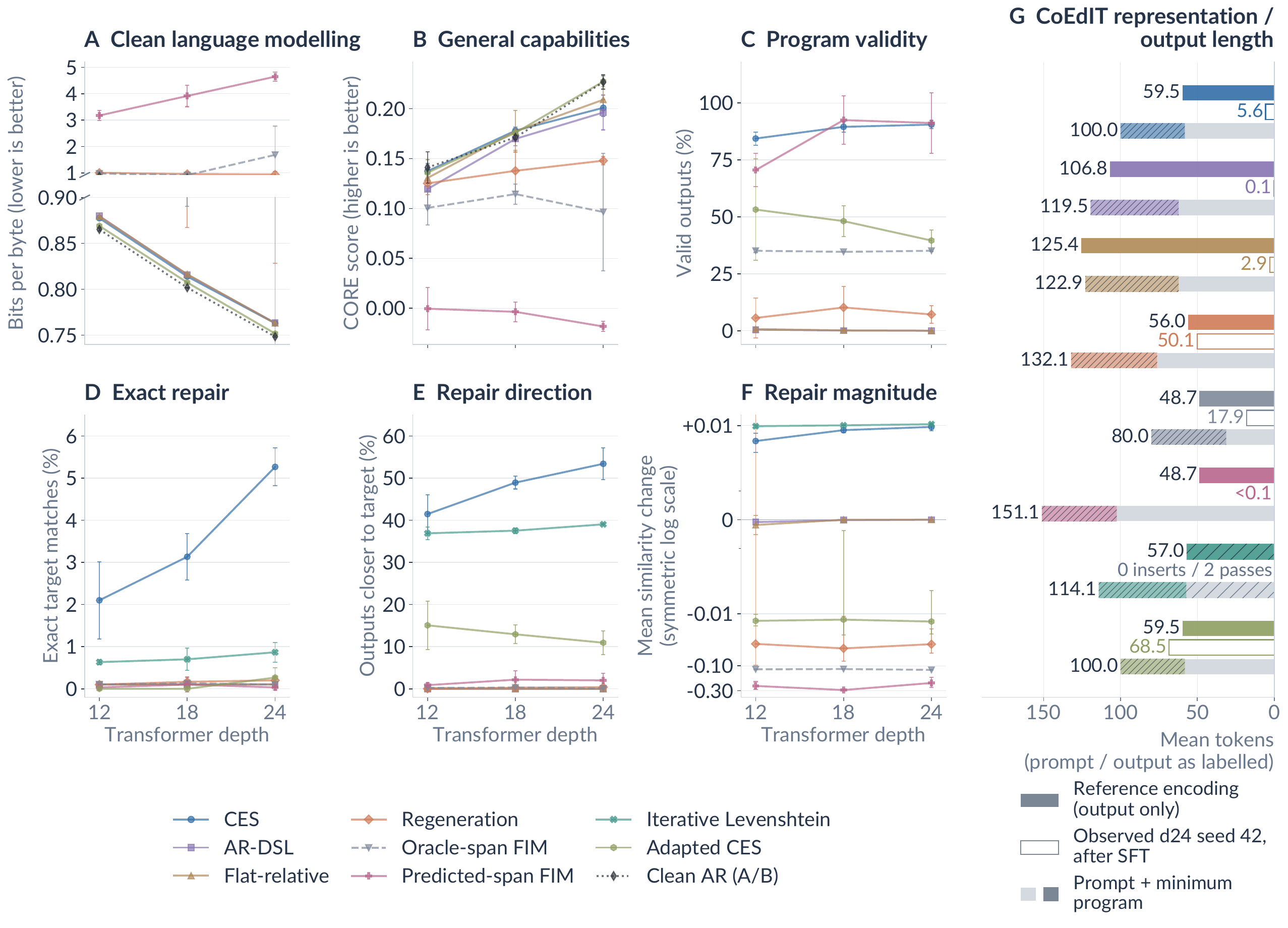}
\caption{\textbf{Complete editing, language modeling and representation cost across methods.}
A/B: clean ClimbMix BPB and CORE performance. C--F: controlled ClimbMix repair.
A--F use pretrained checkpoints, showing means and sample SDs over three
training seeds; repair suites contain 1,000 inputs per run (250 for adapted CES).
C: successful program closure, parsing and application, or text termination
for regeneration/FIM. D: exact target recovery. E/F: frequency and mean magnitude
of character-similarity improvement. Invalid or unterminated outputs return
the source and remain in the denominators; oracle FIM receives gold spans.
Clean AR appears only in A/B.
G: lengths on 485 CoEdIT pairs. Filled bars encode references; outlines count
observed d24 seed-42 outputs after supervised CoEdIT fine-tuning, including
failed attempts. Stacked bars combine prompts (grey) and minimum reference
completions (striped colour), using fixed templates and token boundaries and
excluding COPY/MOVE. The iterative editor's striped reference and prompt bars
measure sequence storage, not autoregressive output. Short observed outputs
need not indicate successful editing; Appendix~\ref{sec:systemscomparison}
defines the accounting.}
\label{fig:methodscale}
\end{figure}

\begin{table}[htbp]
\centering\small
\setlength{\tabcolsep}{4pt}
\renewcommand{\arraystretch}{1.12}
\begin{tabularx}{\textwidth}{@{}>{\raggedright\arraybackslash}p{27mm}>{\raggedright\arraybackslash}X@{}}
\toprule
Method & Mechanism and successful output for \texttt{She walk to school.} $\to$ \texttt{She walks to school.} \\
\midrule
CES & Predicted source addresses and dual RoPE.\newline
$\mathcal{R},\mathcal{R}_{\mathrm S}^{3},\mathcal{R}_{\mathrm E}^{3},\texttt{\textvisiblespace walks},\mathcal{R}_{\mathrm{EOS}},\mathcal{E}_{\mathrm E}$ \\
AR-DSL & Ordinary BPE text encodes operation, numeric coordinates, and payload.\newline
\texttt{R 2 3 " walks"}, then \texttt{</edit-script>} \\
Flat-relative & Dedicated backward-distance vocabulary tokens; ordinary RoPE.\newline
\texttt{R <|back\_0012|> <|back\_0013|> " walks"}, then script closure. \\
Regeneration & Rewrite the complete target, including unchanged text.\newline
\texttt{She walks to school.}, then the text terminator. \\
Oracle-span FIM & Given the target-derived hole between \texttt{She} and \texttt{\textvisiblespace to school.}:\newline
emit \texttt{\textvisiblespace walks}, then the text terminator. \\
Predicted-span FIM & Predict boundaries $(2,3)$ through a pointer head; then fill the span:\newline
emit \texttt{\textvisiblespace walks}, then the text terminator. \\
Iterative\newline Levenshtein & Parallel refinement: delete \texttt{\textvisiblespace walk}, insert one mask after \texttt{She}, then fill it with \texttt{\textvisiblespace walks}. \\
Adapted CES & Add CES components to a clean pretrained decoder; generate the same program as CES above. \\
\bottomrule
\end{tabularx}
\caption{Baseline designs with a shared worked repair, not sampled outputs. Opening controls are supplied; \texttt{\textvisiblespace} marks a leading space. The source includes BOS at index 0: DSL/FIM spans are half-open, so \texttt{walk} occupies $[2,3)$. CES adds a boundary marker and uses inclusive addresses, hence $(3,3)$. Appendix~\ref{sec:baselinespecs} gives full formats. Oracle FIM receives gold locations; adapted CES has an additional adaptation budget.}
\label{tab:baseline_designs}
\end{table}

\subsection{Pretraining Comparison and Representation Cost}
Figure~\ref{fig:methodscale} compares pretrained models and representation costs.
The controlled suite is balanced across insertion, deletion, replacement, movement, and mixed repair, with 1, 2, 4, or 8 operations. The eight-operation stratum is a denser stress condition. These tests measure recovery of known targets, not whether a reference is the only acceptable revision of natural text.

\paragraph{Language modeling and general capabilities (A/B).}
Clean AR has lower validation bits per byte (BPB) than from-scratch CES at all
three depths, while CORE differences vary with depth. BPB measures ordinary
next-token prediction on clean ClimbMix text (lower is better); the centered
CORE aggregate \citep{li2025datacomplmsearchgenerationtraining} measures sampled
language-understanding and knowledge-task performance.
At depth 24, CES obtains 0.7629 BPB versus 0.7484 for clean AR, a relative
increase of approximately 1.9\%.

\paragraph{Complete repair (C/D).}
CES usually produces executable programs, but exact target recovery remains
difficult: depth-24 pretrained CES achieves $90.53\pm1.85$\% valid programs
and $5.27\pm0.45$\% exact recovery. AR-DSL and flat-relative addressing produce no valid controlled
programs in this setting. This supports the combined CES addressing design
under the evaluated recipe; it does not isolate the pointer head from
dual-index RoPE or establish that textual patches generally fail.

\paragraph{Partial repair (E/F).}
CES brings more inputs closer to the target than iterative Levenshtein at
depth 24 ($53.40\pm3.75$\% versus $39.03\pm0.21$\%), but its mean similarity
gain is smaller ($0.00941\pm0.00147$ versus $0.01062\pm0.00046$).
Similarity change is
\[
\Delta S=S(\hat y,y^*)-S(x,y^*),
\]
where $S$ is Python's \texttt{difflib.SequenceMatcher} similarity ratio, based
on matching character blocks (\texttt{autojunk=False}). Panel E reports
$\Pr(\Delta S>0)$; F reports $\mathbb{E}[\Delta S]$, retaining both helpful
and harmful changes. These target-recovery metrics do not measure the
semantic quality of unrestricted revision.

Across the three tested depths, CES improves in clean-language modeling,
CORE, program validity, and both the frequency and mean magnitude of partial
repair; exact recovery also rises, but remains low at 5.27\% at depth 24.

\paragraph{Localization and payload generation.}
Supplying edit locations does not solve content generation. Oracle FIM still
produces harmful payloads, and a separate seed-42 CES diagnostic obtains only
7.3\% depth-24 exact recovery with gold pointers and 8.4\% with the gold
operation envelope as well (Table~\ref{tab:oracleces} in
Appendix~\ref{sec:extendedbaselines}). Predicting locations remains difficult:
predicted-span FIM achieves exact span accuracies of 4.67\%, 3.93\%, and 2.37\%
at the three depths, although its poor BPB also indicates a weak trained comparator.

In a matched single-seed depth-12 ablation, sharing one pointer token across
roles yielded similar language-modeling and editing scores, while synthetic
pointer accuracy fell from 48.64\% to 45.81\%
(Appendix~\ref{sec:sharedpointer}).

\paragraph{Pretrained adaptation.}
Adapted CES retains much of its initial clean-language performance, but its
short adaptation schedule gives weak repair. It establishes compatibility
with pretrained weights, not a compute-matched advantage of adaptation.

\paragraph{Representation cost and preservation (G).}
CES preserves text outside its selected operations and represents sparse
revisions without regenerating unchanged text. On CoEdIT pairs
\citep{raheja-etal-2023-coedit}, minimum insert/delete/replace completions
average 41.8 tokens for CES versus 56.0 for whole-target regeneration;
including prompts gives approximately 100.0 versus 132.1 tokens. These
reference-based costs assume fixed prompts and BPE boundaries; compactness
does not establish edit quality or a quality-matched speedup.

Terminal editing requires no cache rebuild, and generation can continue from the appended source-plus-program history. Continuing instead from materialized edited text requires cache recomputation; its cost depends on context length and rebuild frequency. Full sequence accounting and measured editing and continuation costs are given in Appendix~\ref{sec:systemscomparison}.

Appendix~\ref{sec:extendedbaselines} evaluates supervised CoEdIT fine-tuning using reference-distance metrics and ERRANT (the grammatical Error Annotation Toolkit) \citep{bryant2017errant}, which compares predicted edits with reference edits to compute precision, recall, and $F_{0.5}$. After supervised fine-tuning, regeneration scores higher on reference-based GEC metrics while CES retains positive controlled repair, motivating targeted post-training; the separate judged evaluation below does not establish a reversal of this baseline ranking.

\subsection{Post-Training: Judged Grammatical Correction}
\label{sec:rl}
\begin{figure}[t]
\centering
\includegraphics[width=\textwidth]{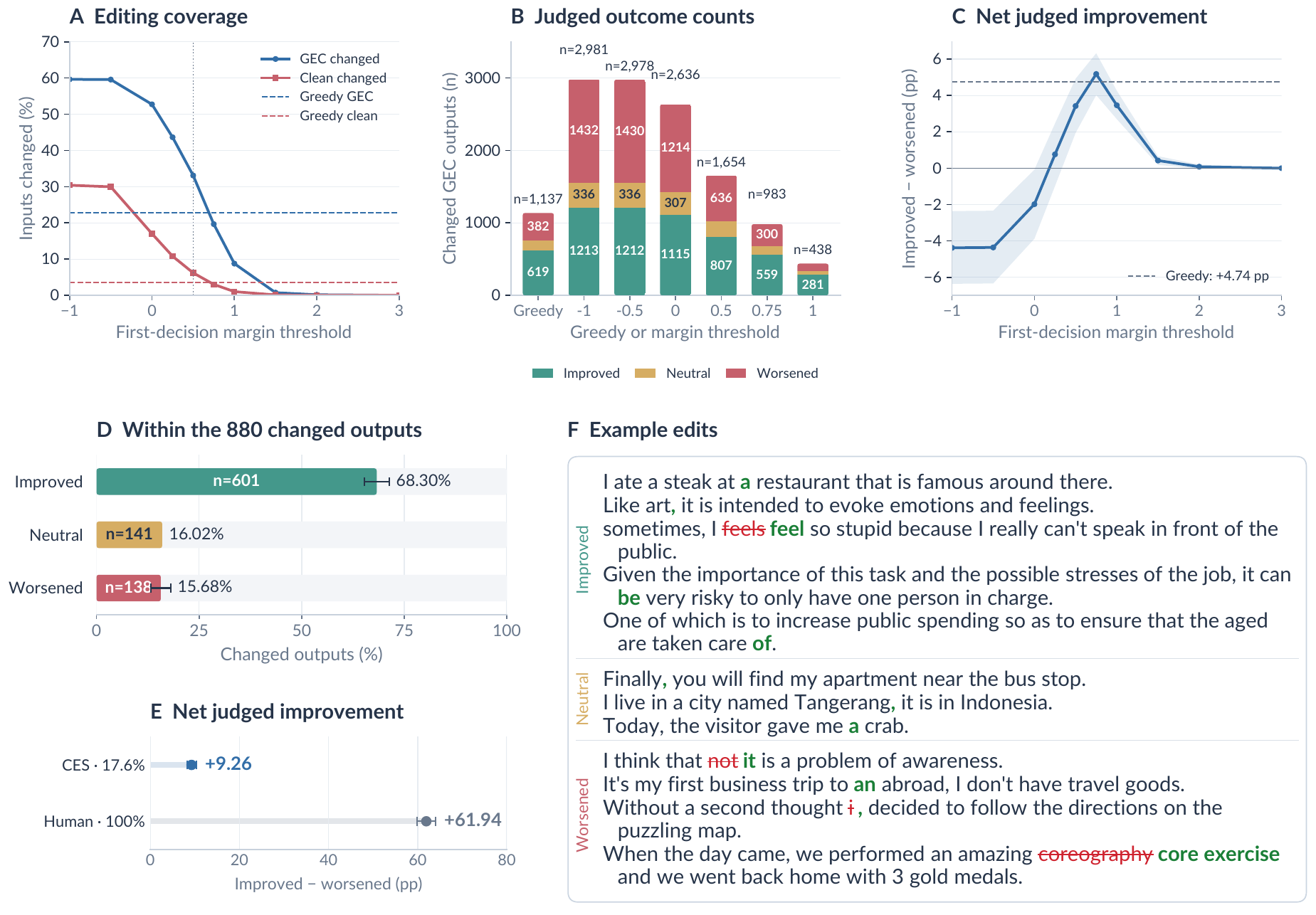}
\caption{\textbf{CES correction with and without post-generation filtering.}
The same d24 checkpoint and 5,000 CoEdIT inputs, plus 5,000 nominal-clean counterparts, underlie both evaluations.
A--C: unfiltered threshold sweep; dashed lines show independently decoded plain greedy and the dotted line marks threshold 0.5. B shows judged counts; C shows all-input improvement minus worsening. All 2,983 distinct changed outputs have reference-blind GPT-5 Mini judgments.
D--E: conservative acceptance policy at threshold 0.5. D conditions on 880 retained changes; E uses all-input denominators and includes the human-reference control. CES and human-reference coverage in E are 17.6\% and 100\%. Intervals are 95\% example bootstraps. Separate judge runs make the filtered/unfiltered contrast descriptive.
F: manually selected plain-greedy examples without a gate or post-generation filters (5 improved, 3 neutral, 4 worsened). Red strike-through marks deletions and green bold marks insertions; indented continuation lines belong to the same example. Category counts are illustrative, not representative frequencies.}
\label{fig:rlpolicy}
\end{figure}

CoEdIT pairs source sentences with grammatical corrections \citep{raheja-etal-2023-coedit},
but acceptable revisions can differ from the reference. The selected d24 seed-42
checkpoint is therefore assessed by reference-blind GPT-5 Mini judging after the
curriculum in Section~\ref{sec:gecposttraining}. Figure~\ref{fig:rlpolicy} compares
unfiltered decoding and a conservative acceptance policy, reporting improvement,
harm, coverage, and nominal-clean preservation together.

\paragraph{Evaluation and decoding policy.}
A custom 5,000-pair CoEdIT confirmation partition is used, with its recorded targets supplied as 5,000 nominal-clean identity inputs. The judge sees the source and anonymous model/human candidates, without identifying the human edit as an answer key. Appendix~\ref{sec:rlsplits} describes the partition and source-overlap checks.
An optional threshold $\tau$ controls whether editing begins: unlike plain
greedy, the gate combines probabilities across edit openers, so $\tau=0$
is not equivalent to greedy decoding (Appendix~\ref{sec:rlrecipe}).

\paragraph{Unfiltered edits and selectivity.}
Plain greedy yields positive net judged improvement of $+4.74$ percentage
points (95\% CI [3.52,5.96]), changing 22.74\% of inputs: 619 improve,
136 are neutral and 382 worsen. It changes 3.48\% of nominal-clean inputs.
More aggressive editing increases harm: at $\tau=-1$, coverage reaches
59.62\% but net improvement falls to $-4.38$ points. At $\tau=0.75$, net
improvement is $+5.18$ points with 19.66\% coverage; this threshold is
exploratory rather than independently validated. Appendix~\ref{sec:rlunfiltered}
gives decoding limits and replay checks; Figure~\ref{fig:rlpolicy}F illustrates
successes, neutral changes, and failures.

\paragraph{Conservative acceptance policy.}
Filtering yields more selective positive results. At $\tau=0.5$, candidates
are accepted only if they change at most four removed-plus-inserted characters
across at most one region and have mean program-decision log probability at least $-0.6$;
rejections return the source. At 17.60\% coverage, 68.30\% of retained changes
improve and 15.68\% worsen, giving all-input net improvement of $+9.26$ points.
Only 1.48\% of nominal-clean inputs change. The human control has 100\%
changed coverage and $+61.94$ points, so it is not coverage-matched.
Together, the unfiltered and filtered results establish positive judged editing
capability in a CES model, but do not isolate the contribution of RL or
establish superiority to general GEC systems;
full-curriculum/filter-matched controls and independent human judging remain absent.

\subsection{Continuation After Edit Programs}
\label{sec:continuationmain}
CES can also condition subsequent predictions on an appended edit program
while retaining the original source history. On 540 held-out ClimbMix
documents, a source prefix is corrupted and a correct repair program is
supplied. The next 128 tokens of the original document are then scored under
teacher forcing. Relative to the same corrupted prefix followed by an empty
program, the repair lowers mean negative log likelihood (NLL) by 0.02123
nats/token. The reduction is 0.00641 nats/token relative to a valid sham program
with the same operation family and span size that leaves the prefix incorrect.
Thus the supplied correction makes the original continuation more likely,
providing evidence that the model uses the repair content. This tests
interpretation of supplied repairs, not autonomous repair or sampled
continuation quality. Appendix~\ref{sec:posteditcontinuation}
gives the protocol and uncertainty; Appendix~\ref{sec:systemscomparison}
reports cache-rebuilding costs when continuing from materialized edited text.

\section{Conclusion}
CES integrates explicit source addressing and edit programs into a causal decoder.
The experiments establish controlled repair, beneficial post-trained edits under
model judging, and continuation conditioned on supplied repairs. These results
support causal referencing as a practical modeling interface; reliable unrestricted
revision and transfer to larger pretrained models remain open.

\section{Limitations and Future Work}
\label{sec:limitations}
\paragraph{Coverage and generalization.}
Experiments use one NanoChat model family up to 1.389B parameters and a 2,048-token context. Exact repair remains low, dense edits are difficult, and insertion recovery is a persistent weakness. Generated training edits are replacement-heavy and copy/move coverage is uneven. The trained regeneration, FIM, and iterative baselines do not represent state-of-the-art editing systems. Transfer to modern LLM families remains untested.

\paragraph{Post-training and measurement.}
The selected confirmation policy permits only small local changes and abstains on 82.4\% of editing inputs. Its positive semantic direction is specific to a GPT-5 Mini judge, candidate panel, custom holdout, and one post-training seed. The same model family generated some supervision and judged the outputs. Changing the set of candidates presented to the judge changed scores for identical CES outputs, and pair-based splitting allowed 24 repeated sources. Example bootstraps do not capture training or judge uncertainty. There is no independent blinded human audit or confirmation-set comparison with continued SFT or regeneration matched for the full training curriculum and acceptance filters.

\paragraph{Systems and next steps.}
Appending programs preserves caches but does not reclaim context; materializing
edited text requires rebuilding. Timings depend on hardware, kernels, and context.
Source-grouped splits, coverage-matched comparisons with stronger editors,
and broader copy/move and long-context evaluations are needed.

\paragraph{Future work.}
A shared pointer token could support more general source references whose roles
are specified through instructions or in-context examples, extending the ablation
in Appendix~\ref{sec:sharedpointer}. Improved initialization and adaptation training
for the localization head and dual-index attention could help integrate these
mechanisms into pretrained LLMs. Joint training on programs and pointers could
teach models to generate code that reads referenced spans and writes computed
results back to selected locations. Better teacher repairs and task distributions
could develop strong referencing with less corruption training. Equal-FLOP
comparisons and scaling studies could examine how model capacity and corruption
exposure affect editing ability and ordinary autoregressive performance. Larger
models and practical text, code, and agent-context tasks would test these
capabilities beyond controlled repair.

References could remain unexpanded until the model requests materialization of
a region for editing, with addresses remapped to the expanded content. This makes
cache reuse, faster partial rebuilding, and approximate KV-cache updates important
systems directions. An edit-based discrete diffusion model is another possibility:
training across noise levels could support generation from random text through
repeatedly generated and applied edit programs, with the aim of approximating
the training distribution.

\section*{Impact Statement}
Explicit edit programs can make text and code revision easier to inspect. They can also support subtle document manipulation or introduce harmful changes into otherwise correct text. Structural validity and model confidence are insufficient safeguards for semantic correctness. The current models are research artifacts; downstream applications should expose proposed changes for review and preserve provenance. Generated data and derivatives retain applicable upstream restrictions.

\section*{Software and Data}
CES extends NanoChat \citep{nanochat}. Code is available at
\url{https://github.com/Louiii/CES}; data, tokenizer, and model checkpoints
are available at \url{https://huggingface.co/datasets/lorob/CES}.

\label{page:mainend}

\clearpage
\renewcommand{\acksection}{\section*{Acknowledgements and Disclosures}}
\begin{ack}
\paragraph{Funding.}
This research received no external funding and was conducted independently of the author's employment.

\paragraph{Competing interests.}
The author is employed by InstaDeep. This work was conducted independently and does not represent the views of InstaDeep.
\end{ack}

\bibliographystyle{plainnat}
\bibliography{references}

\clearpage
\appendix
\section{Reproducibility, Protocols, and Asset Provenance}
\label{sec:reproducibility}
The comparison contains eight editing methods at three depths and three training seeds; each is evaluated before and after supervised CoEdIT fine-tuning. Nine clean AR checkpoints are additionally reported as language-model references.

\subsection{Training budgets and implementation}
Table~\ref{tab:trainingbudgets} gives training budgets and optimization settings.
\begin{table}[tbp]
\centering\small
\begin{tabular}{lrrr}
\toprule
Setting & d12 & d18 & d24 \\
\midrule
CES parameters (millions) & 287.44 & 704.55 & 1,388.84 \\
Causal pretraining updates & 26,880 & 39,600 & 74,800 \\
Serialized token slots (millions) & 880.80 & 1,297.61 & 2,451.05 \\
Iterative editor updates & 8,960 & 13,200 & 24,933 \\
Clean AR reference updates & 26,880 & 39,600 & 4,675 \\
Additional CES adaptation updates & 1,000 & 1,000 & 1,000 \\
Supervised causal GEC updates & 150 & 150 & 150 \\
Iterative GEC updates & 50 & 50 & 50 \\
\bottomrule
\end{tabular}
\par\medskip
\setlength{\tabcolsep}{4pt}
\begin{tabular}{lrrrr}
\toprule
Pretraining stage & Slots/update & LR warmup & LR decay fraction & Muon WD schedule \\
\midrule
Six causal editing methods & 32,768 & 40 & 0.20 & Constant \\
Additional CES adaptation & 32,768 & 40 & 0.20 & Constant$^*$ \\
Iterative Levenshtein & 32,768 & 40 & 0.20 & Constant, unscaled \\
\bottomrule
\end{tabular}
\caption{Training budgets and optimization settings for the method comparison. Context length is 2,048. Nominal token-slot budgets and clean-token exposure are distinct quantities. Warmup is in updates; LR decay occupies the final stated fraction and ends at 0.05 of the peak rate. $^*$Adaptation weight decay is disabled during its initial frozen-backbone stage. Iterative Levenshtein uses three encoder passes per step; pass-count matching approximates computation, not equal FLOPs or wall-clock cost.}
\label{tab:trainingbudgets}
\end{table}

Training uses Muon for matrix parameters and AdamW for embeddings, unembedding, and scalar parameters. Editing-method pretraining uses H100 tensorwise FP8 training and evaluation in the base compute dtype. Model/data seed pairs are $(42,1234)$, $(43,1235)$, and $(44,1236)$ for the causal methods; the iterative trainer uses seeds 42/43/44 for both initialization and data generation. The final position-loss weight is 0.01 for CES, adapted CES, and predicted FIM.

\paragraph{Shared causal pretraining settings.}
The raw matrix/embedding/unembedding/scalar learning-rate arguments for the causal grid are 0.02/0.3/0.008/0.5; the trainer applies batch and width scaling. The nominal Muon weight decay is 0.28, scaled by batch size and model parameter counts.

\paragraph{Iterative training.}
The iterative trainer applies the same batch and width learning-rate scaling, but its Muon weight decay remains 0.28 without the causal trainer's scaling. Both trainers ramp Muon momentum from 0.85 to 0.97 over 400 updates; only the causal trainer decreases it to 0.90 during the final LR decay.

\paragraph{Supervised CoEdIT fine-tuning.}
All eight editing methods are separately fine-tuned on the 12,756 CoEdIT source--target training pairs, without mixing in clean-text replay. This supervised comparison is distinct from the balanced SFT and RL curriculum in Appendix~\ref{sec:rlrecipe}. The seven causal methods use 16,384 token slots per update, raw matrix/embedding/unembedding/scalar learning rates 0.002/0.03/0.0008/0.05, a nominal constant Muon weight decay of 0.05, 20 warmup steps, a 0.30 warmdown fraction, and a final learning-rate fraction of 0.10. The same batch/width LR and Muon decay scaling are applied. The iterative trainer uses 50 three-pass updates rather than 150 causal updates, 10 warmup steps, and unscaled Muon weight decay 0.05.

\paragraph{Pretrained-decoder adaptation.}
Of its 1,000 additional updates, the first 250 train only new components;
imported parameters then use a 0.1 learning-rate multiplier. The additional
adaptation budget and separate clean pretraining prevent an equal-budget
comparison with from-scratch CES. Architecture and initialization are
specified in Appendix~\ref{sec:baselinespecs}.

\subsection{Metric and evaluation contracts}
Controlled repair uses 1,000 examples before fine-tuning and a 250-example
stratified subset afterward; method-grid GEC metrics use the 485 official
\CoEdit{} validation examples. CES pointer accuracy, textual-coordinate
accuracy, and exact FIM span accuracy measure different decisions and are
not compared as a shared metric.

\paragraph{Clean-language evaluation.}
BPB uses 1,048,576 validation-token positions, sequence length 2,048, and
batches of eight. CORE uses a fixed selection of up to 50 examples per task
with few-shot contexts. Clean AR is evaluated only on these language-model
metrics.

\paragraph{Editing metrics.}
Character similarity uses \texttt{SequenceMatcher}, including for iterative
Levenshtein; GEC closer/farther rates use token Levenshtein distance, and
semantic improvement/harm uses the stated model judge. Invalid and rejected
outputs resolve to the source in unconditional metrics. Table~\ref{tab:oracleces}
reports oracle CES diagnostics; Appendix~\ref{sec:sharedpointer} reports
shared-pointer accuracy.

\subsection{Compute resources}
At depth 24, pretraining on one H100 (80GB) takes approximately 8 hours for CES,
AR-DSL, regeneration, and predicted-span FIM; 9--10 hours for flat-relative and
oracle-span FIM; 6.7 hours for adapted CES (including clean pretraining); and
10.6 hours for iterative Levenshtein. For d24 CES on two RTX 5080s (16GB each),
the 800-update balanced SFT stage takes approximately 21 minutes, followed by
33 minutes of broad RL (100 updates) and 41 minutes of GEC RL (200 updates).
These are recorded training-loop times, excluding setup, checkpoint I/O, and evaluation.

The shared-pointer ablation uses RTX 5080s.
The cache-rebuilding benchmark uses an RTX 5080; editing-latency comparisons
use an H100 (Appendix~\ref{sec:systemscomparison}). GPT-5 Mini pretraining-edit generation
used approximately 0.5B API tokens and \$200 in credits, excluding
GEC corruption filtering, semantic judging, and post-training experiments.

Main confirmation judging cost approximately \$5; auxiliary judging and
corruption-filtering experiments cost approximately \$40, estimated from recorded
GPT-5 Mini API usage.

Recorded method-comparison training logs account for approximately 260 H100
GPU-hours and 34 RTX 5080 GPU-hours. These totals cover recorded completed
training segments, excluding the post-training stages above, evaluation, and
additional preliminary or unsuccessful experiments. The unfiltered confirmation
inference on 5,000 editing and 5,000 clean inputs takes approximately 10 minutes
on one RTX 5080, including the plain-greedy and forced-edit decodes.

\FloatBarrier
\section{Corruption Distribution and Data Quality}
\label{sec:corruption}

Self-supervised training pairs are generated by sampling a stochastic stream
of insertions, deletions, replacements, and moves under a fixed serialization
budget. The process is parameterized by edit-spacing and edit-length
distributions, an insertion-content mixture, and optional protection of a
clean prefix as stable context.

\paragraph{Edit arrival and lengths.}
Distances to candidate deletion and insertion events follow geometric
distributions, with per-token start probabilities $p^{\text{del}}_{\text{start}}$
and $p^{\text{ins}}_{\text{start}}$:
\[
\Delta^{\text{del}} \sim \mathrm{Geom}(p^{\text{del}}_{\text{start}}), \qquad
\Delta^{\text{ins}} \sim \mathrm{Geom}(p^{\text{ins}}_{\text{start}}).
\]
The earlier event determines the next operation. Edit lengths follow
\[
L^{\text{del}} \sim \mathrm{Geom}(p^{\text{del}}_{\text{end}}), \qquad
L^{\text{ins}} \sim \mathrm{Geom}(p^{\text{ins}}_{\text{end}}),
\]
where $\mathrm{Geom}(p)$ returns a positive integer with mean $1/p$. In practice these are parameterized via means $(\mu^{\text{del}}_{\text{gap}}, \mu^{\text{del}}_{\text{len}}, \mu^{\text{ins}}_{\text{gap}}, \mu^{\text{ins}}_{\text{len}})$ by setting $p = 1/\mu$.

\paragraph{Token-budgeted serialization.}
Sampling stops when the next edit would exceed the serialized token budget;
that edit is dropped. Operation counts vary within the fixed maximum length.

\paragraph{Insert-content source mixture.}
For each insertion-style payload of length $L^{\text{ins}}$, inserted tokens are drawn from a mixture distribution over sources:
\[
s \sim \mathrm{Categorical}(\pi_{\text{buf}}, \pi_{\text{self}}, \pi_{\text{uni}}, \pi_{\text{marg}}),
\]
The sources are a replay reservoir (\textsc{Buffer}), a span from the current
sequence (\textsc{Self}), uniform random tokens (\textsc{Uniform}), and a
marginal unigram distribution (\textsc{Marginal}). Reported random-corruption
runs use weights 0.6/0.1/0.2/0.1, renormalized over available sources, with
uniform sampling as fallback when no non-uniform source is available.

\paragraph{Protecting a prefix as stable context.}
A sampled protection length $P$ restricts edits to the final $N-P$ tokens,
leaving the first $P$ tokens as uncorrupted conditioning context.

\paragraph{Source text and generated edits.}
ClimbMix is the 400B-token pretraining mixture associated with Nemotron-CLIMB
\citep{diao2025climb}. An approximately 15GB shard subset of NanoChat's shuffled release,
\texttt{karpathy/climbmix-400b-shuffle}, is used. The 2.5\% sample used for GPT-5 Mini edit generation is drawn from this
subset, not the full 400B-token corpus. The bootstrap reserves
\texttt{shard\_06542.parquet} for validation; the source corpus is distributed
under CC BY-NC 4.0. GPT-5 Mini proposes localized edits, which are parsed into
the CES language. These are synthetic revisions, not human annotations.
GPT-5 Mini emits one operation per line in a text-based domain-specific
language: \texttt{Del(region)}, \texttt{Replace(region, text)},
\texttt{Insert(pointer, text)}, or \texttt{Move(region, pointer)}.
A \texttt{RegionInclusive} identifies a source span using exact search phrases
for its start and end, with \texttt{start\_exact} and \texttt{end\_exact}
substrings marking the inclusive boundaries within those phrases.
A \texttt{Pointer} identifies an insertion boundary using a source search
phrase and adjacent \texttt{pre} and \texttt{post} strings: insertion occurs
between them. For example, \texttt{pre="chicken, "} and
\texttt{post="seafood"} identify the boundary in \texttt{chicken, seafood}.
All anchors refer to the unmodified source; only the \texttt{text} payload
may introduce new content. During data preparation, these text anchors are
resolved against the source and converted to token-coordinate edits for CES
serialization.

\begin{figure}[tbp]
\centering\includegraphics[width=\textwidth]{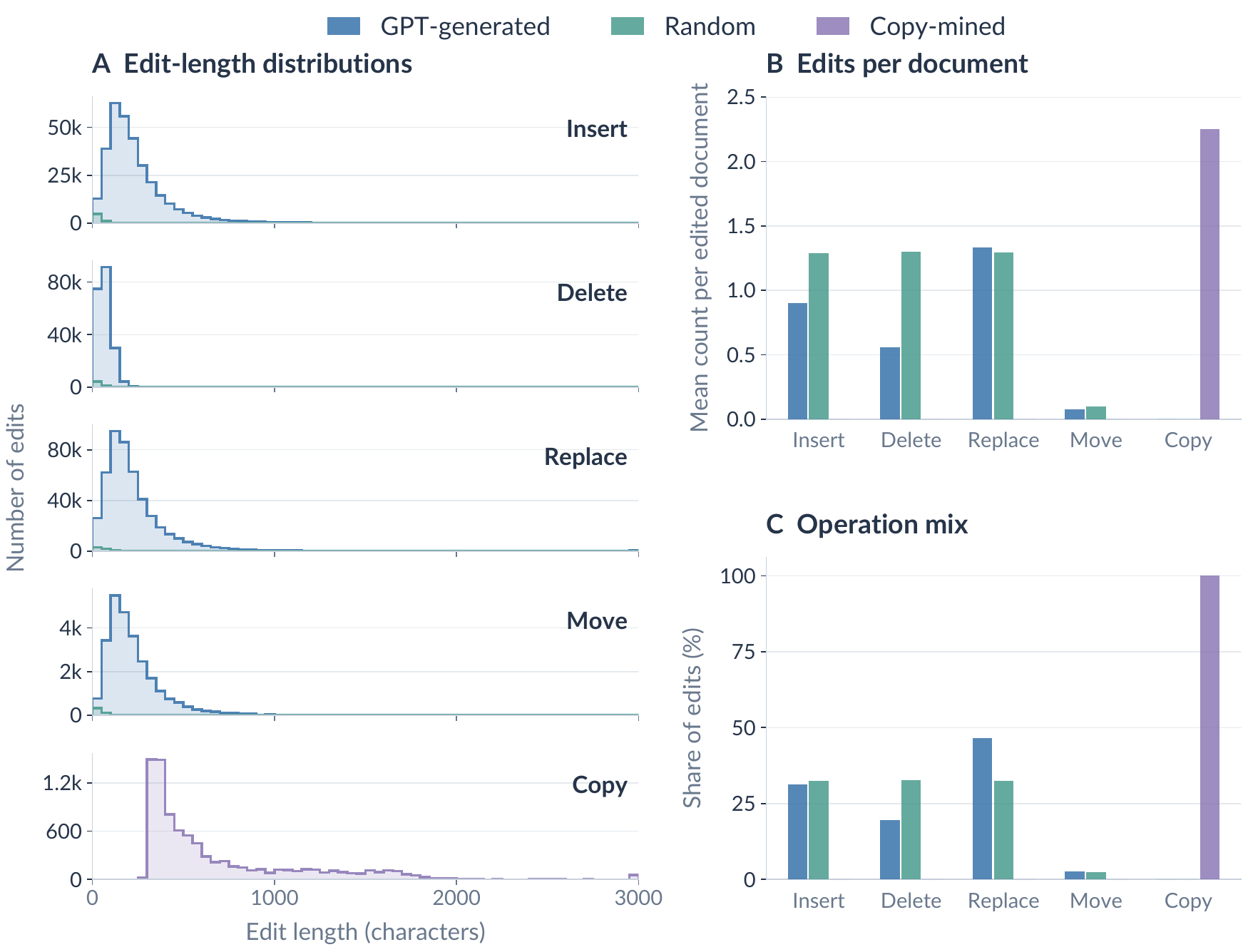}
\caption{Operation and edit-length distributions in GPT-generated, random, and copy-mined training data. GPT-generated edits are replacement-heavy and longer than random edits; copy mining supplies long repeated spans. Appendix~\ref{sec:supervision} compares training with these sources.}
\label{fig:dataset_stats}
\end{figure}

\paragraph{Copy mining and source differences.}
A separate pipeline removes later occurrences of repeated spans and creates
copy programs that restore them from earlier occurrences. These examples
complement the GPT-5 Mini and random-corruption sources
(Figure~\ref{fig:dataset_stats}).

\subsection{Structural and semantic quality}
\label{sec:dataquality}
The edit corpus produced by GPT-5 Mini for pretraining contains 361,773 rows,
of which 361,429 pass structural checks. Its approximately 1.03M operations
comprise 201k insertions, 325k deletions, 482k replacements, and 26k moves.
Parsing and round-trip resolution verify serialization, not semantic quality
or human preference.

For GEC post-training, filtering proceeds from 12,000 prepared clean targets
to 11,895 audit items, 11,887 returned judgments (eight failures), 6,847
semantic-audit passes, 6,813 after localization/distance filtering, and 6,245
after held-out-source exclusion. Unlike the pretraining structural checks,
this audit assesses meaning using an automated judge; it does not replace
human validation. Appendix~\ref{sec:rlrecipe} specifies its use in post-training.

\FloatBarrier
\section{Edit Decoding and Cache Management}
\label{sec:inference_details}

Algorithm~\ref{alg:ces} generates a program against frozen source coordinates,
with grammar masks restricting token roles and legal addresses. Invalid or
unfinished programs return the source and remain in evaluation denominators;
generation limits and acceptance policies are specified with each evaluation.

\begin{algorithm}[H]
\caption{Causal edit decoding with explicit prefill}
\label{alg:ces}
\renewcommand{\algorithmiccomment}[1]{\hfill\textit{\textcolor{CESMuted}{#1}}}
\newcommand{\CESModel}{\textcolor[HTML]{326EA8}{\mathop{\text{\textnormal{CESGPT}}}}}
\newcommand{\CESTokenLogits}{\textcolor[HTML]{1A7F37}{v_{\mathrm{logits}}}}
\newcommand{\CESAddressLogits}{\textcolor[HTML]{765C9E}{p_{\mathrm{logits}}}}
\newcommand{\CESV}{\textcolor[HTML]{1A7F37}{v}}
\newcommand{\CESP}{\textcolor[HTML]{765C9E}{p}}
\begin{algorithmic}[1]
\REQUIRE Tokenized source prompt $x$, address indices $p_x$, budget $B$
\STATE $\_,\_,c \gets \CESModel(x,\text{\textnormal{address}}=p_x,\text{\textnormal{cache}}=\varnothing)$ \COMMENT{prefill}
\STATE $\CESV \gets \mathcal{E}_{\text{S}}$; $\CESP \gets |x|$; $\text{\textnormal{program}} \gets []$.
\FOR{$t=1,\ldots,B$}
  \STATE $\CESTokenLogits,\CESAddressLogits,c \gets \CESModel([\CESV],\text{\textnormal{address}}=[\CESP],\text{\textnormal{cache}}=c)$ \COMMENT{decode}
  \STATE Sample legal token $\CESV$ from $\CESTokenLogits$.
  \STATE Set $\CESP$ to a legal address selected from $\CESAddressLogits$ if $\CESV$ is a pointer;
  \STATE \hspace{1em}otherwise use the edit location for a payload, or the stream index.
  \STATE $\text{\textnormal{program.append}}((\CESV,\CESP))$
  \IF{$\CESV=\mathcal{E}_{\text{E}}$}
    \STATE Apply $\text{\textnormal{program}}$ to the source in $x$ and return the result; return the source on failure.
  \ENDIF
\ENDFOR
\STATE Return the source if the budget is exhausted before program completion.
\end{algorithmic}
\end{algorithm}

The prompt $x$ includes the boundary marker and preceding context but excludes
edit-start, whose zero-based position is $|x|$. After prefilling cache $c$,
each iteration processes one token/address pair; the accumulated program
supports grammar masking and edit application.

\paragraph{Terminal editing.}
Applying the program returns revised text without rebuilding the source cache.

\paragraph{Continuation from the serialized history.}
Generation resumes after $\mathcal{E}_{\text{E}}$, retaining the source and
program. Appending tokens preserves earlier KV entries but does not compact
the history. Section~\ref{sec:continuationmain} tests whether the model can
interpret supplied correct programs in this mode.

\paragraph{Continuation from materialized text.}
The revised text replaces the conditioning prefix. States from the earliest
changed token or position must be recomputed; earlier states can be reused.
Appendix~\ref{sec:systemscomparison} measures full-prefix rebuilds.

A full edit--continue--return interface could repeatedly generate and apply
programs, update token/address streams, and resume from materialized text
with the affected cache rebuilt, until a dedicated return token ends the
interaction. This requires training the return decision and was not evaluated.

\FloatBarrier
\section{Supervision Ablations}\label{sec:supervision}

\begin{figure}[tbp]
\centering
\includegraphics[width=0.94\textwidth]{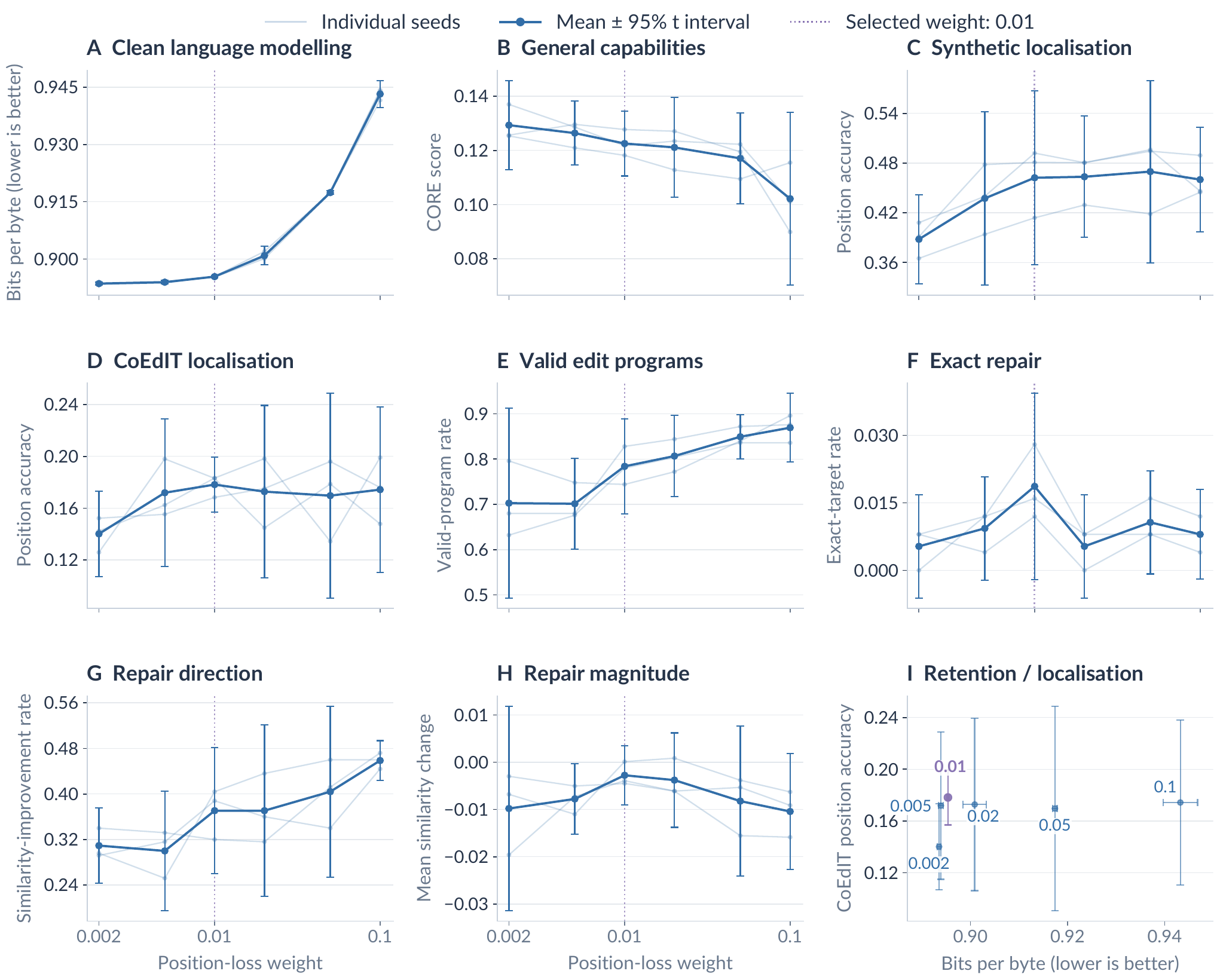}
\caption{Position-head weight selection over three paired depth-12 seeds.
The selected $\lambda_{\mathrm{pos}}=0.01$ is the retention/editing Pareto knee. Error bars are two-sided 95\% Student-$t$ intervals over training seeds; faint lines show individual seeds.}
\label{fig:positionweights}
\end{figure}

\paragraph{Position-loss weight sweep.}
The depth-12 position-weight sweep (Figure~\ref{fig:positionweights}) selects
$\lambda_{\mathrm{pos}}=0.01$ as the observed Pareto knee. It obtains most of
the localization and controlled-validity gain before clean-language
degradation becomes steep. Relative to $0.005$, validity rises by 8.27
percentage points and controlled similarity-improvement rate by 7.07 points,
while validation \BPB\ changes by only 0.16\% relative.

\begin{figure}[tbp]
\centering
\includegraphics[width=0.94\textwidth]{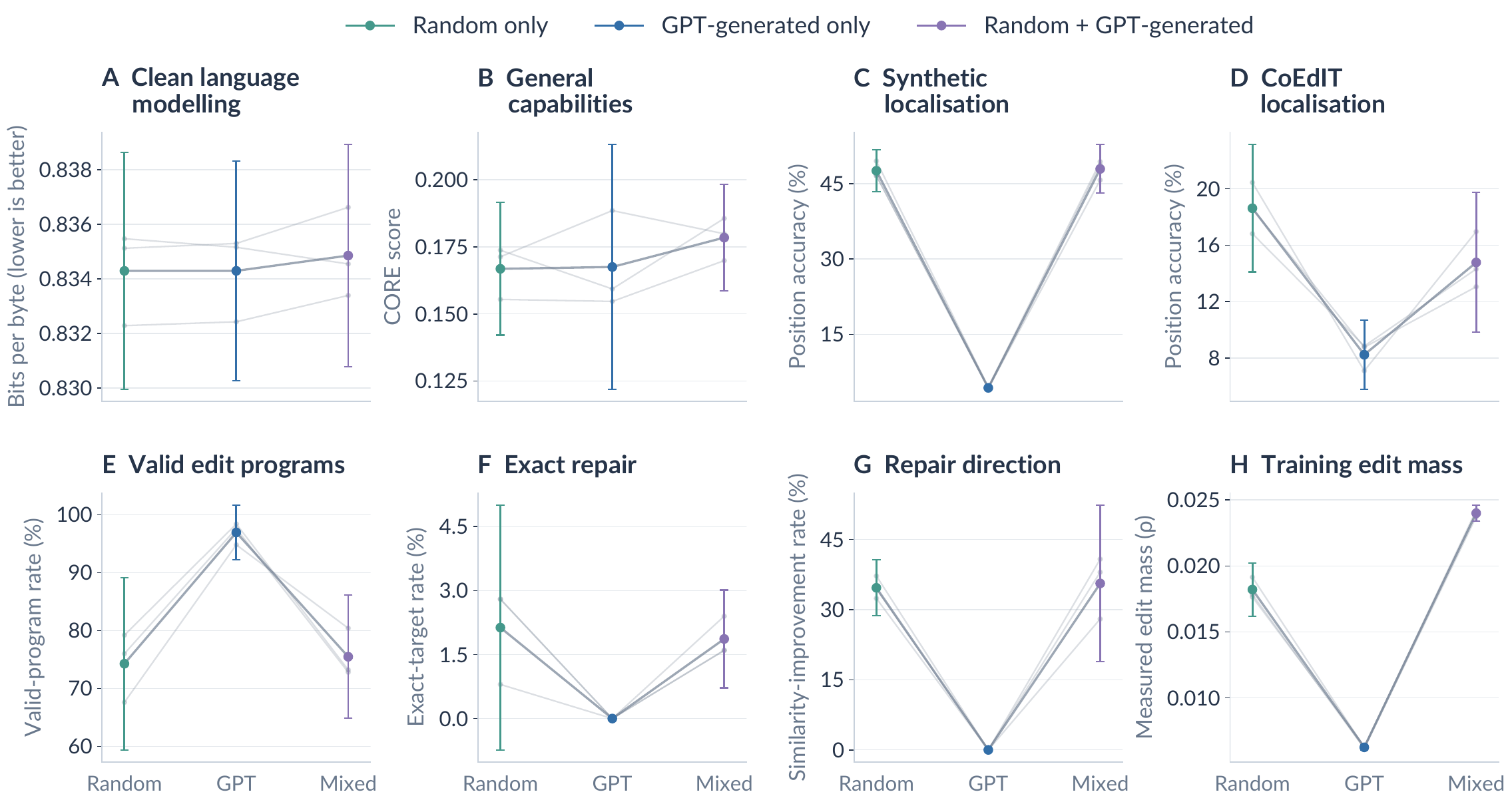}
\caption{Dataset-source comparison over three paired depth-18 seeds. Validity
must be read together with source copying and nontrivial editing. Error bars are 95\% Student-$t$ intervals over the three paired seeds.}
\label{fig:datasetsources}
\end{figure}

\paragraph{Training-data source sweep.}
At depth 18, mixed random and GPT-5 Mini data (Figure~\ref{fig:datasetsources}) obtains the best CORE score
(0.17846) and valid nontrivial-edit rate (50.93\%). Random-only remains
competitive and is better on human localization, exact recovery, and mean
character direction. GPT-5 Mini-only is not competitive: 96.93\% syntactic
validity is dominated by 90.40\% unchanged output. Random fallback is therefore
essential, while mixed data is the pragmatic source choice.

\paragraph{Corruption-exposure sweep.}
Measured training edit mass is
\[
\rho=\frac{\text{total script edit mass}}{\text{total raw source tokens consumed}}.
\]
Mass counts insertion payload tokens, deleted span tokens, both the removed
span and new payload for replacement, and the source span once for copy or
move. It is computed from the supplied scripts, excluding operation and
pointer tokens. The corruption control adjusts training exposure; $\rho$
measures the resulting token ratio. Control $1.0$ yields mean
$\rho\approx0.02358$ in this sweep.

\begin{figure}[tbp]
\centering
\includegraphics[width=0.92\textwidth]{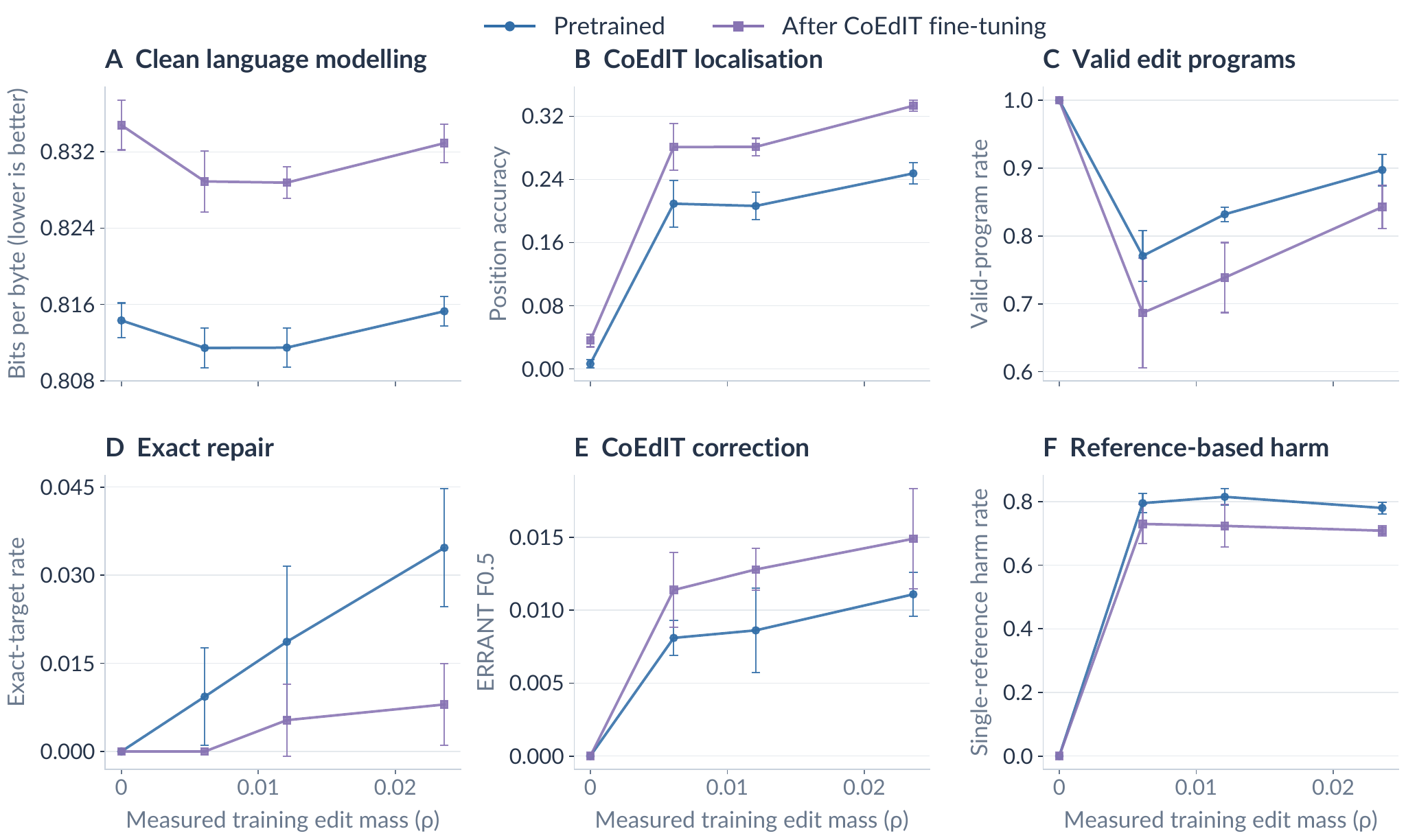}
\caption{Retention and transfer as a function of measured edit mass $\rho$,
before and after \CoEdit\ fine-tuning. Points show means and error bars one sample standard deviation over three training seeds.}
\label{fig:rhotradeoffs}
\end{figure}

The corruption-exposure sweep (Figure~\ref{fig:rhotradeoffs}) selects control $1.0$ as a multi-metric Pareto
choice, not a statistically unique optimum. It provides the strongest human
localization, controlled validity and exact recovery, and post-fine-tuning
\ERRANT\ $F_{0.5}$, at a modest clean-retention cost. With only three seeds,
nearby \ERRANT\ differences are not decisive.

\FloatBarrier
\section{Decoding-policy sensitivity}
\label{sec:samplingpolicy}

We compare greedy decoding with temperatures $T\in\{0.2,0.5,0.8,1.0\}$,
nucleus sampling with $p\in\{0.90,0.95\}$, and top-$k$ sampling with
$k\in\{20,50\}$ on 1,000 controlled repairs and 485 \CoEdit{} examples.
Each stochastic setting uses decoding seeds 12345--12347 from the same
seed-42 trained checkpoint; greedy is evaluated once. Iterative Levenshtein
is deterministic and has only a greedy row.

\begin{table}[tbp]
\centering
\small\setlength{\tabcolsep}{3pt}
\begin{tabular}{lrrrrrr}
\toprule
Policy & Valid & Exact & $\Delta S$ & $F_{0.5}$ & Closer & Farther \\
\midrule
Greedy & 0.8870 & \textbf{0.0570} & \textbf{+0.00877} & 0.01990 & 0.0536 & 0.7031 \\
$T=0.2$ & 0.8960 & 0.0543 & +0.00818 & 0.01934 & 0.0605 & 0.6907 \\
$T=0.5$ & 0.9163 & 0.0440 & +0.00643 & \textbf{0.02184} & \textbf{0.0770} & \textbf{0.6660} \\
$T=0.8$ & 0.9163 & 0.0363 & +0.00271 & 0.02153 & 0.0674 & 0.6701 \\
$T=1.0$ & 0.9100 & 0.0307 & -0.00121 & 0.01631 & 0.0488 & 0.7079 \\
Nucleus $p=0.90$ & \textbf{0.9180} & 0.0357 & +0.00221 & 0.01923 & 0.0591 & 0.6873 \\
Nucleus $p=0.95$ & 0.9100 & 0.0327 & +0.00070 & 0.01797 & 0.0529 & 0.7017 \\
Top-$k$, $k=20$ & 0.9120 & 0.0313 & +0.00215 & 0.01801 & 0.0543 & 0.6797 \\
Top-$k$, $k=50$ & 0.9083 & 0.0310 & +0.00070 & 0.01837 & 0.0536 & 0.6893 \\
\bottomrule
\end{tabular}
\caption{Decoding sensitivity for the depth-24 seed-42 \CES{} checkpoint.
Stochastic rows are means over three decoding seeds; the deterministic greedy
row is a single evaluation. Bold marks the best value in each column.}
\label{tab:cessampling}
\end{table}

\begin{figure}[tbp]
\centering
\includegraphics[width=0.98\textwidth]{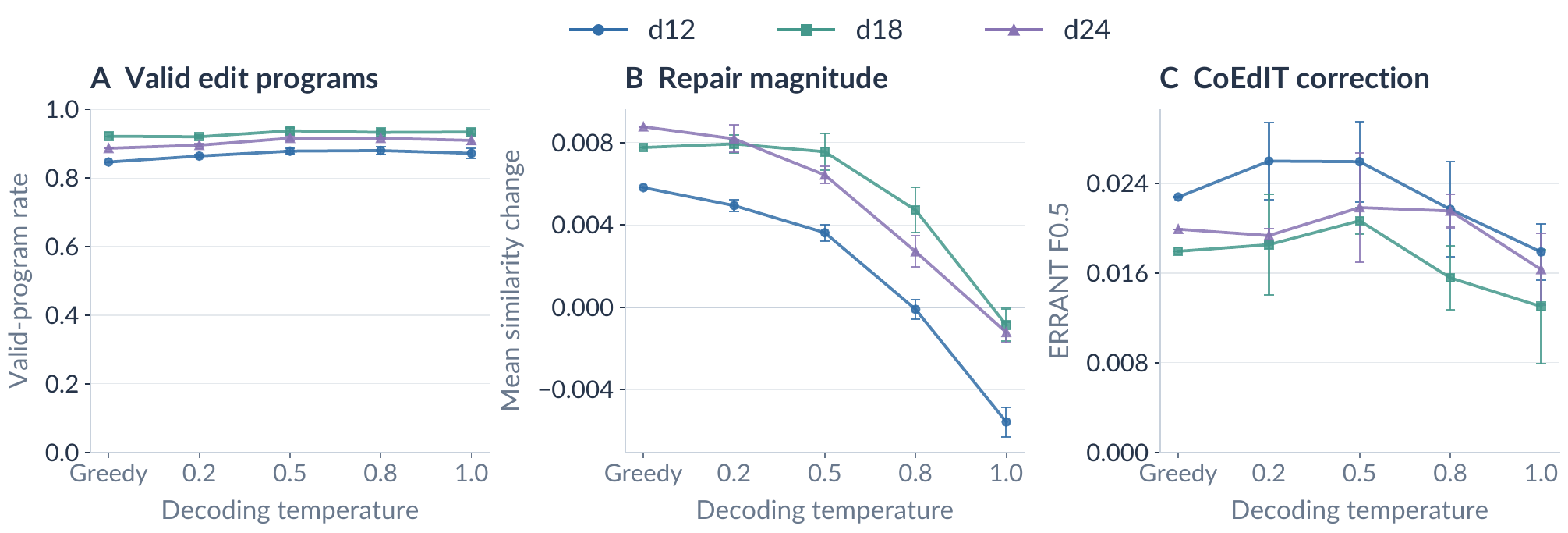}
\caption{CES temperature sensitivity across model depth. Error bars for
stochastic settings are one sample standard deviation over three decoding
seeds; greedy is deterministic.}
\label{fig:cessamplingdepth}
\end{figure}

For depth-24 \CES{}, moderate temperature improves validity and reference-based
GEC metrics at the cost of exact recovery and similarity gain
(Table~\ref{tab:cessampling}). At $T=1.0$, controlled similarity gain becomes
negative. No stochastic policy dominates greedy; moderate temperatures
preserve the main scaling conclusions across depths
(Figure~\ref{fig:cessamplingdepth}).

\begin{figure}[tbp]
\centering
\includegraphics[width=\textwidth]{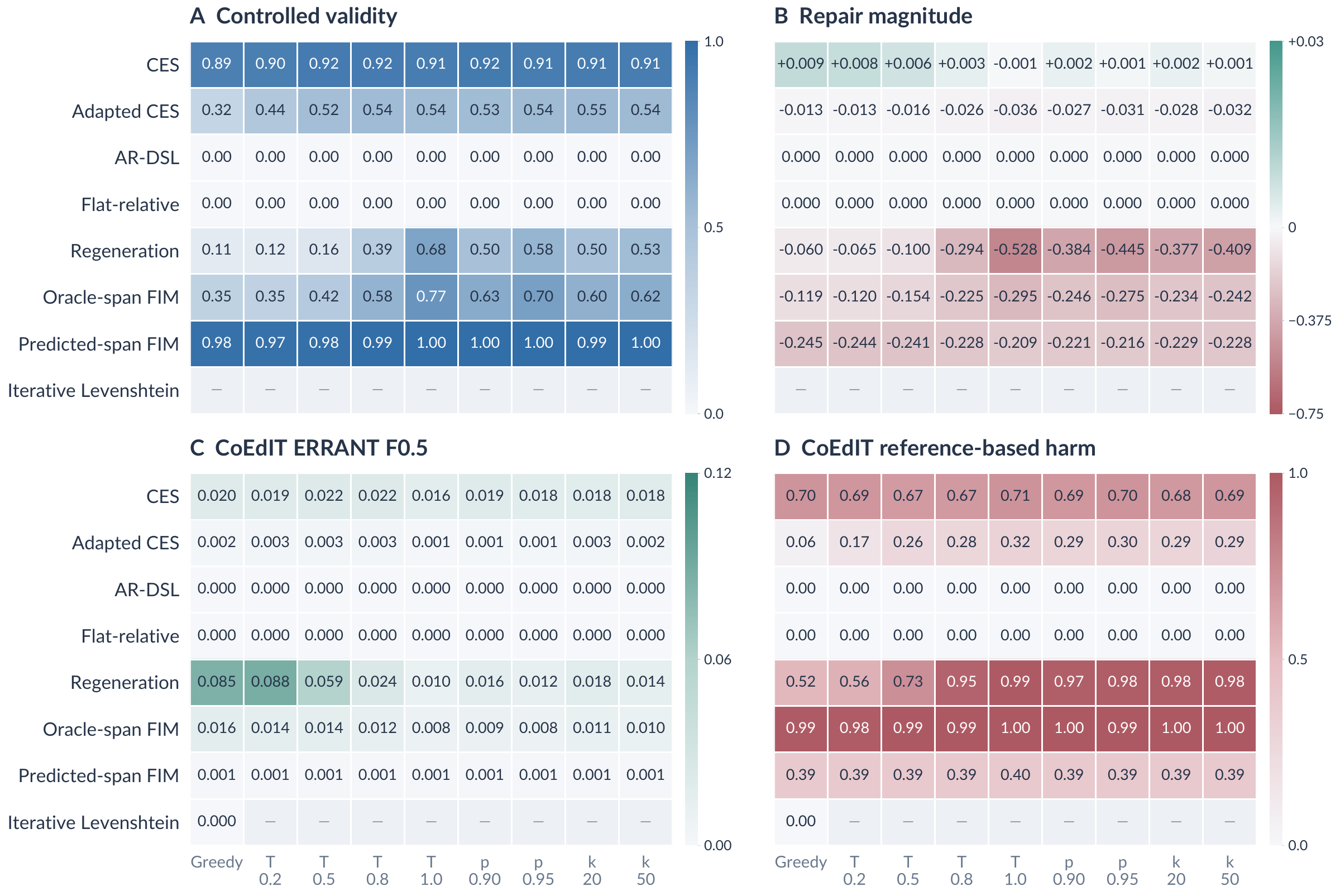}
\caption{Depth-24 policy sensitivity for all methods. Each stochastic cell is
the mean of three decoding seeds from one trained checkpoint. A dash marks a
not-applicable or unevaluated cell. Zero harm for methods that copy the source must
not be read as useful safety; it accompanies zero correction coverage.}
\label{fig:samplingdepth24}
\end{figure}

Sampling does not reverse the qualitative baseline conclusions
(Figure~\ref{fig:samplingdepth24}). Regeneration and oracle-span \FIM{}
terminate more often at higher temperatures but produce worse edits.
\ARDSL{} and flat-relative remain ineffective across policies; adapted
\CES{} gains validity while increasing harm, and predicted-span \FIM{}
retains negative controlled similarity gain despite frequent termination.

\FloatBarrier
\section{Baseline Architectures and Edit Interfaces}
\label{sec:baselinespecs}

Causal decoders use width $64$ times depth, full-context causal attention, and the NanoChat tokenizer. The methods map corrupted source text $x$ to revised text $\hat y$ using different prediction targets and reconstruction procedures. Table~\ref{tab:baseline_designs} illustrates their outputs for the same repair; training settings are given in Appendix~\ref{sec:reproducibility}.

\subsection{Shared conventions}
Let $x=(x_0,\ldots,x_{n-1})$, with document BOS at $x_0$. Textual DSLs and FIM use half-open source spans $[i,j)$ and insertion boundaries before token $i$. CES instead addresses token positions in the serialized stream with inclusive span ends: source boundary $j$ maps to the stream position of $x_{j-1}$. Coordinates remain fixed throughout a program.

Controlled evaluation uses greedy decoding with budgets of 128 new tokens for CES and 640 for the other causal editing methods, clipped to remaining context. Invalid or incomplete outputs return the source. Method-specific completion rules are given below. Except for predicted FIM and CES, the causal baselines use ordinary RoPE without an address head; predicted FIM adds a head but retains ordinary RoPE.

\subsection{CES and pretrained-decoder adaptation}
CES uses split dual-index RoPE, a source-address head, and the training objective in Section~\ref{sec:trainingobjective}. The evaluated grammar closes replacement payloads with $\mathcal{R}_{\mathrm{EOS}}$; insertions end at the next operation opener or edit-end, without a separate insertion terminator. From-scratch CES splits each Q/K head equally between stream and address
rotations without changing its width. Adapted CES copies a clean AR
backbone and shared vocabulary rows, retaining full-width stream Q/K
projections and appending full-width address projections. The rotated,
normalized branches are concatenated; a zero-initialized gate on the address
query preserves the initial attention logits. Scaling remains $h_d^{-1/2}$
for original head width $h_d$; values are zero-padded and extra output
channels discarded. The edit interface and loss are unchanged; the adaptation
schedule is specified in Appendix~\ref{sec:reproducibility}.

The inference prompt supplies the source, editable-boundary marker, and edit-start. Token and address decisions follow native grammar and legal-address masks until edit-end. No additional payload-length or operation-count cap is imposed in this comparison. The completed program is applied to the original source.

\subsection{AR-DSL: numeric addresses in ordinary text}
An edit-bearing training region contains the source, a newline-delimited script enclosed by \texttt{<edit-script>} and \texttt{</edit-script>}, and the ordinary assistant-end token. The grammar is
\[
\begin{array}{ll}
\texttt{D }i\ j & \text{delete }[i,j),\\
\texttt{I }i\ \texttt{"payload"} & \text{insert before }i,\\
\texttt{R }i\ j\ \texttt{"payload"} & \text{replace }[i,j),\\
\texttt{C }i\ j\ \texttt{-> }k & \text{copy }[i,j)\text{ before }k,\\
\texttt{M }i\ j\ \texttt{-> }k & \text{move }[i,j)\text{ before }k.
\end{array}
\]
Addresses are decimal strings tokenized by ordinary BPE; payloads are JSON strings preserving whitespace and quotation marks. The ordered loader interleaves clean and edit-bearing regions. Source and script tokens receive vocabulary supervision, without an auxiliary coordinate loss.

Inference starts from the source followed by a newline and \texttt{<edit-script>}, without a DSL-specific grammar mask. Generation uses ordinary decoder stopping rules. The evaluator extracts the first closed script, validates its coordinates and overlap constraints, and applies it; validity requires a closed, applicable script.

\subsection{Flat-relative: addresses as backward-distance tokens}
Flat-relative shares AR-DSL's training layout, operation syntax, payloads, prompt, and parser. Each numeric coordinate is replaced by a dedicated vocabulary token $\texttt{BACK}_{r-i}$, where $r$ is the coordinate token's serialized index and $i$ the source boundary. Available distances are 1--2,047. References to the same boundary can therefore require different tokens at different stream positions.

Distance tokens receive vocabulary supervision and are converted to source coordinates before parsing; out-of-range addresses are invalid. This comparison changes the whole CES address mechanism, including both RoPE and the localization head, rather than ablating only the head.

\subsection{Whole-target regeneration}
Regeneration uses the following prefix, preceded by BOS, for pretraining, supervised CoEdIT fine-tuning, and editing evaluation:
\begin{quote}\small\ttfamily
Rewrite the corrupted text as clean text.\\
Corrupted text:\\
\normalfont\emph{source text}\\
\ttfamily Clean text:
\end{quote}
Training appends the entire target and assistant-end token. Only this completion receives vocabulary loss; the instruction and source are conditioning inputs. The pair loader uses resolved source/target pairs, including identity pairs. Clean-language BPB and CORE evaluation do not use this editing template.

The model regenerates unchanged as well as changed text. Decoding stops on assistant-end or document BOS, which is stripped from the output. The evaluator recognizes termination by stopping before budget exhaustion; exhausted budgets or overlong prompts return the source. Validity denotes a terminated completion, not a correct repair.

\subsection{Oracle-span fill-in-the-middle}
Removing shared outer tokens from source and target defines the smallest enclosing changed region. The prompt contains BOS, \texttt{<fim-prefix>}, retained prefix, \texttt{<fim-suffix>}, retained suffix, and \texttt{<fim-middle>}, with newlines around each cue. The cues are ordinary tokenized text. Loss covers only the target middle and assistant-end token.

Inference receives this target-derived envelope, making oracle FIM information-advantaged: localization is supplied, but payload content must be generated. One envelope may include unchanged text between separated edits. Decoding follows regeneration's budget and stopping rules; reconstruction joins the retained prefix, generated middle, and retained suffix.

\subsection{Predicted-span fill-in-the-middle}
Predicted FIM adds a localization head. Two query states, following the source at \texttt{<fim-locate-start>} and \texttt{<fim-locate-end>}, score source tokens. Boundary $i$ is represented by token $i-1$, allowing empty insertion spans and an end-of-source boundary. Training supervises these queries with the target-derived envelope and address-loss weight 0.01, then appends the FIM prompt and target middle. Vocabulary loss covers the middle and assistant-end token.

At inference, the start is the highest-scoring source position and the end the highest-scoring position at or after it; adding one converts each to a boundary. The predicted FIM prompt is appended to the cached source/locator history without removing or prefilling the source again. Infill decoding follows regeneration's stopping rules, with its budget clipped to the combined prompt's remaining context. Reconstruction uses the predicted span; overlong prompts return the source. Both localization and generation are evaluated, within a single enclosing span.

\subsection{Iterative Levenshtein editing}
The iterative baseline is depth-matched, without reproducing every training choice of the original Levenshtein Transformer. Its bidirectional transformer with ordinary RoPE predicts token deletions, insertion counts per gap, and inserted-token identities. It has no causal next-token interface or address head. Aligned source/target pairs express all edits, including replacement, copy, and move, as deletions and insertions.

Three encoder passes train the heads with cross-entropy: the current sequence supplies deletion labels, the oracle-kept sequence gap counts, and the oracle-masked sequence inserted-token labels. Counts range from 0 to 32 per gap; longer insertions span multiple oracle rounds. The model uses private BOS, EOS, mask, and padding tokens.

Inference alternates deletion, gap-count prediction, and mask filling for at most four rounds, protecting BOS/EOS. A round with no edits stops refinement; one without insertions skips filling. The current sequence is returned at the round limit. The evaluator strips leaked control tokens and records their frequency. Autoregressive BPB, decoder-based CORE, and causal-program validity do not apply. Inference work is measured by pass and inserted-token counts rather than serialized output length.

\subsection{Clean autoregressive reference}
Clean AR is trained on uncorrupted ClimbMix through next-token prediction. It has no trained edit prompt and appears only in Figure~\ref{fig:methodscale}A/B.

\paragraph{Comparison scope.}
Shared corpora and nominal token budgets do not imply identical supervision or numbers of source examples. These comparisons evaluate complete editing systems rather than isolated architectural changes.

\FloatBarrier
\FloatBarrier
\section{Extended baseline and human-transfer results}
\label{sec:extendedbaselines}
Tables~\ref{tab:d24pretrain}--\ref{tab:d24gec} give the depth-24 values and single-reference GEC metrics supporting Figure~\ref{fig:methodscale}.
For the supervised CoEdIT fine-tuning comparison, changed coverage is $\Pr(\hat y\ne x)$. Closer/farther rates compare token Levenshtein distances to the recorded target: $\Pr[d(\hat y,y^*)<d(x,y^*)]$ and $\Pr[d(\hat y,y^*)>d(x,y^*)]$. ERRANT $F_{0.5}$ weights precision more strongly than recall \citep{bryant2017errant}. These reference-based metrics can penalize acceptable alternative revisions and differ from the GPT-5 Mini judgments in Section~\ref{sec:rl}.

Regeneration has the highest depth-24 mean ERRANT score and reference-improvement rate after supervised CoEdIT fine-tuning, despite strongly negative controlled similarity change. CES retains positive controlled repair but moves farther from the single human reference on $68.45\pm2.70$\% of CoEdIT inputs. This transfer gap motivates the separate post-training application; it should not be read as a comparison of the final RL policy against these SFT baselines.

\begin{table}[t]
\centering\small
\setlength{\tabcolsep}{3pt}
\begin{tabular}{lrrrrr}
\toprule
Method & BPB & CORE & Valid (\%) & Exact (\%) & Copy (\%) \\
\midrule
Clean AR & $0.7484\pm0.0006$ & $0.227\pm0.007$ & -- & -- & -- \\
CES & $0.7629\pm0.0015$ & $0.201\pm0.009$ & $90.53\pm1.85$ & $5.27\pm0.45$ & $9.53\pm1.85$ \\
AR-DSL & $0.7634\pm0.0004$ & $0.196\pm0.017$ & $0.00\pm0.00$ & $0.10\pm0.00$ & $100.00\pm0.00$ \\
Regeneration & $0.9365\pm0.1079$ & $0.148\pm0.004$ & $7.13\pm3.86$ & $0.20\pm0.10$ & $92.93\pm3.88$ \\
Oracle FIM$^\dagger$ & $1.6816\pm1.0851$ & $0.096\pm0.059$ & $35.10\pm0.44$ & $0.10\pm0.00$ & $64.90\pm0.44$ \\
Predicted FIM & $4.6474\pm0.1787$ & $-0.018\pm0.005$ & $91.23\pm13.23$ & $0.03\pm0.06$ & $16.67\pm10.48$ \\
Flat-relative & $0.7630\pm0.0006$ & $0.209\pm0.011$ & $0.00\pm0.00$ & $0.10\pm0.00$ & $100.00\pm0.00$ \\
Iterative Levenshtein & -- & -- & -- & $0.87\pm0.23$ & $58.07\pm0.55$ \\
Adapted CES & $0.7517\pm0.0006$ & $0.227\pm0.005$ & $39.60\pm4.61$ & $0.27\pm0.23$ & $60.53\pm4.81$ \\
\bottomrule
\end{tabular}
\caption{Depth-24 pretrained results, means and sample SDs over three training seeds. BPB is measured on validation text. Controlled metrics use 1,000 examples per seed, except adapted CES (250). Copy includes invalid-as-source fallback. $^\dagger$Oracle FIM receives the reference span. Clean AR uses the same nominal token budget; no edit capability is evaluated. Iterative Levenshtein has no comparable causal BPB or program-validity diagnostic.}
\label{tab:d24pretrain}
\end{table}

\begin{table}[t]
\centering\small
\setlength{\tabcolsep}{3pt}
\begin{tabular}{lrrrrr}
\toprule
Method & Changed (\%) & $F_{0.5}$ & Closer (\%) & Farther (\%) & Ctrl. $\Delta S$ \\
\midrule
CES & $90.0\pm2.6$ & $0.0155\pm0.0051$ & $5.4\pm1.2$ & $68.5\pm2.7$ & $0.0092\pm0.0035$ \\
AR-DSL & $0.0\pm0.0$ & $0.0000\pm0.0000$ & $0.0\pm0.0$ & $0.0\pm0.0$ & $0.0000\pm0.0000$ \\
Regeneration & $72.9\pm2.2$ & $0.0733\pm0.0235$ & $15.3\pm3.8$ & $53.9\pm6.9$ & $-0.6819\pm0.0370$ \\
Oracle FIM$^\dagger$ & $97.9\pm1.0$ & $0.0148\pm0.0011$ & $0.4\pm0.4$ & $97.3\pm1.3$ & $-0.2228\pm0.0062$ \\
Predicted FIM & $37.0\pm12.2$ & $0.0009\pm0.0004$ & $0.8\pm0.3$ & $34.8\pm12.5$ & $-0.0919\pm0.0685$ \\
Flat-relative & $0.1\pm0.2$ & $0.0000\pm0.0000$ & $0.0\pm0.0$ & $0.1\pm0.1$ & $0.0000\pm0.0000$ \\
Iterative Levenshtein & $0.7\pm0.6$ & $0.0000\pm0.0000$ & $0.0\pm0.0$ & $0.5\pm0.4$ & $0.0112\pm0.0004$ \\
Adapted CES & $10.1\pm7.4$ & $0.0017\pm0.0010$ & $0.1\pm0.1$ & $8.4\pm6.2$ & $-0.0000\pm0.0001$ \\
\bottomrule
\end{tabular}
\caption{After supervised CoEdIT fine-tuning, three training seeds. Changed, ERRANT, and reference-distance outcomes use 485 GEC pairs; controlled direction uses a stratified 250-example subset. $F_{0.5}$ is on a 0--1 scale. Oracle FIM is information-advantaged. Zero harm with zero changes represents abstention, not effective correction.}
\label{tab:d24gec}
\end{table}

\FloatBarrier
\subsection{CES with supplied edit locations}
\label{sec:localization}
The single-seed Oracle CES diagnostic separates localization from payload generation (Table~\ref{tab:oracleces}). The gold-pointer
protocol supplies target-derived source coordinates while letting the model
choose operation and payload tokens. The full oracle-span protocol additionally
forces the gold operation envelope. Neither exposes target payload tokens.

\begin{table}[htbp]
\centering
\small\setlength{\tabcolsep}{3pt}
\begin{tabular}{rlrrrrrr}
\toprule
Depth & Oracle input & Pre. valid & Pre. exact & FT $F_{0.5}$ & Coverage & Closer & Farther \\
\midrule
12 & Gold pointers & 0.288 & 0.049 & 0.0303 & 0.214 & 0.078 & 0.039 \\
12 & Gold span + ops & 0.994 & 0.077 & 0.0460 & 0.726 & 0.045 & 0.631 \\
18 & Gold pointers & 0.337 & 0.052 & 0.0275 & 0.231 & 0.087 & 0.052 \\
18 & Gold span + ops & 0.996 & 0.083 & 0.0591 & 0.635 & 0.062 & 0.520 \\
24 & Gold pointers & 0.399 & 0.073 & 0.0245 & 0.223 & 0.076 & 0.052 \\
24 & Gold span + ops & 0.994 & 0.084 & 0.0981 & 0.868 & 0.074 & 0.740 \\
\bottomrule
\end{tabular}
\caption{Single-seed target-aware Oracle CES diagnostics. These protocols are
not deployable end-to-end systems and should not be ranked with the main table.}
\label{tab:oracleces}
\end{table}

Gold pointers produce conservative outputs: depth-24 coverage is 0.223 and
harm 0.052. Forcing the full operation envelope raises depth-24 \ERRANT\
$F_{0.5}$ to 0.0981 and coverage to 0.868, but harm rises to 0.740. Localization
is therefore a real bottleneck, but removing it does not make payload
generation safe.

\FloatBarrier
\section{Autoregressive continuation after an edit program}
\label{sec:posteditcontinuation}

These tests measure whether a supplied correct edit program improves subsequent
AR prediction while retaining the original source history. They evaluate use
of an oracle repair, not autonomous correction.

\subsection{Held-out natural-text likelihood}

We score the same 128-token clean tail after different histories of a
384-token source prefix, using 540 unique held-out ClimbMix validation
documents. The design crosses three operations (replacement, deletion,
insertion), span lengths (1, 4, 16 tokens), and edit-to-tail distances
(16, 64, 128 clean tokens), with 20 documents per cell.

For tail tokens $y_{i,1:T_i}$ and condition-specific history $h_{i,c}$, the
mean tail negative log likelihood is
\begin{equation}
  \overline{L}_{c} =
  \frac{1}{\sum_i T_i}
  \sum_i\sum_{t=1}^{T_i}
  -\log p_\theta\!\left(y_{i,t}\mid h_{i,c},y_{i,<t}\right).
\end{equation}
BPB divides summed NLL by $\log 2$ and the number of UTF-8 tail bytes.
Histories comprise clean/corrupt plain prefixes, clean/corrupt prefixes with
empty programs, and corrupt prefixes with either a gold repair or a valid
non-correcting sham. The sham matches the repair's operation family and span
size but yields an incorrect prefix, controlling for a non-empty program.

Repair gain is defined as
$\overline{L}_{\mathrm{corrupt\ empty}}-
 \overline{L}_{\mathrm{repair}}$, so positive values indicate a more likely
clean tail. Aggregate recovery is
\begin{equation}
  \frac{\overline{L}_{\mathrm{corrupt\ empty}}-
        \overline{L}_{\mathrm{repair}}}
       {\overline{L}_{\mathrm{corrupt\ empty}}-
        \overline{L}_{\mathrm{clean\ empty}}}.
\end{equation}
This ratio is computed from dataset-level losses rather than by averaging unstable
per-example ratios. Confidence intervals use 20,000 paired bootstrap
resamples over documents.

\begin{table}[tbp]
\centering\small
\begin{tabular}{lrr}
\toprule
Metric & Adapted CES (1,000 steps) & Mature CES (74,800 steps) \\
\midrule
Clean empty BPB & 0.8215 & 0.8681 \\
Corrupt empty BPB & 0.8241 & 0.8706 \\
Sham BPB & 0.8250 & 0.8661 \\
Repair BPB & 0.8242 & 0.8642 \\
Gain vs. empty [95\% CI] & $-0.44\ [-2.50,1.80]$ & $+21.23\ [18.68,24.22]$ \\
Gain vs. sham [95\% CI] & $+2.69\ [0.99,4.77]$ & $+6.41\ [3.91,9.56]$ \\
Aggregate recovery & $-5.1\%$ & $257.4\%$ \\
Repair wins vs. empty / sham & 51.5\% / 49.3\% & 90.4\% / 57.8\% \\
\bottomrule
\end{tabular}
\caption{Held-out natural-text continuation. Gains are millinats per tail token with paired 95\% document-bootstrap intervals. The mature checkpoint lowers NLL relative to both empty and non-correcting sham programs. Recovery above 100\% is not an accuracy; see the interpretation in the text.}
\label{tab:posteditnatural}
\end{table}

\begin{figure}[tbp]
\centering
\includegraphics[width=0.98\textwidth]{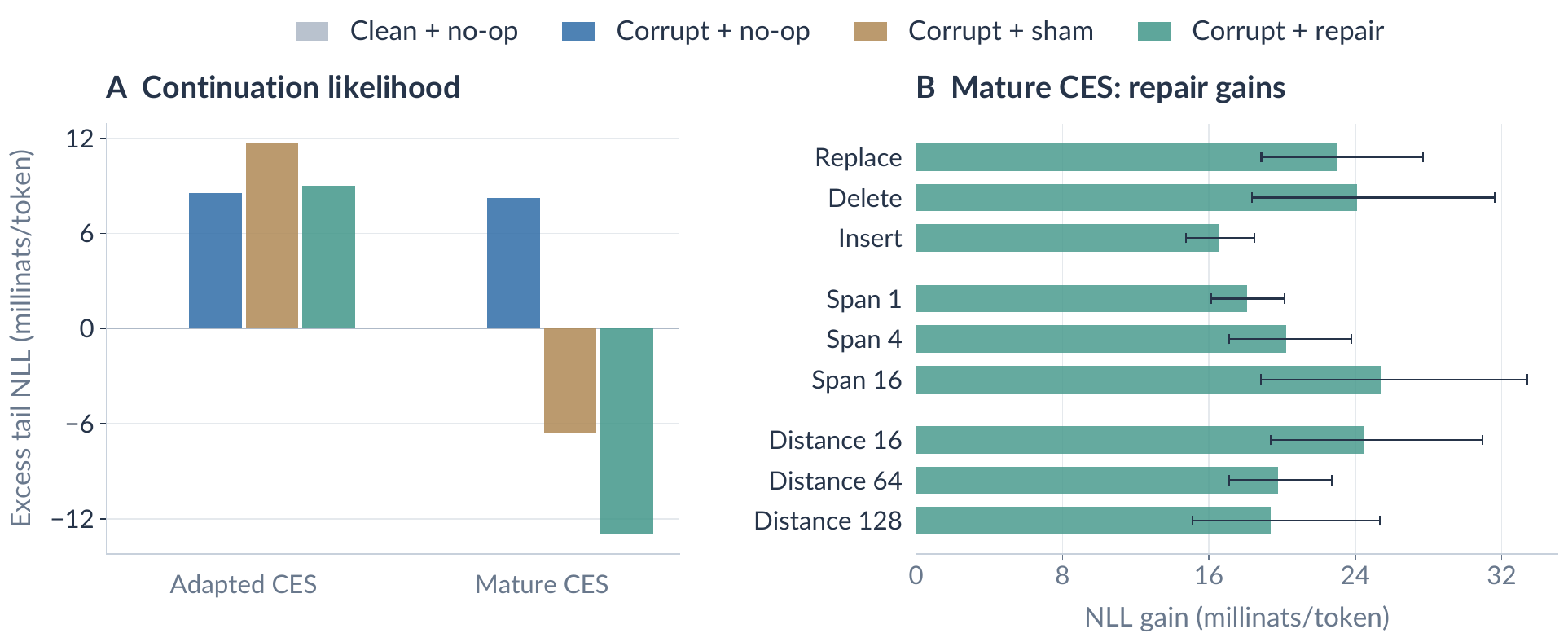}
\caption{Natural-text continuation likelihood. Left: excess clean-tail NLL
relative to clean plus empty program. Right: mature-\CES\ repair gain for every
operation, span, and distance aggregate; error bars are paired 95\% bootstrap
intervals.}
\label{fig:postedittailoverview}
\end{figure}

\begin{figure}[tbp]
\centering
\includegraphics[width=0.98\textwidth]{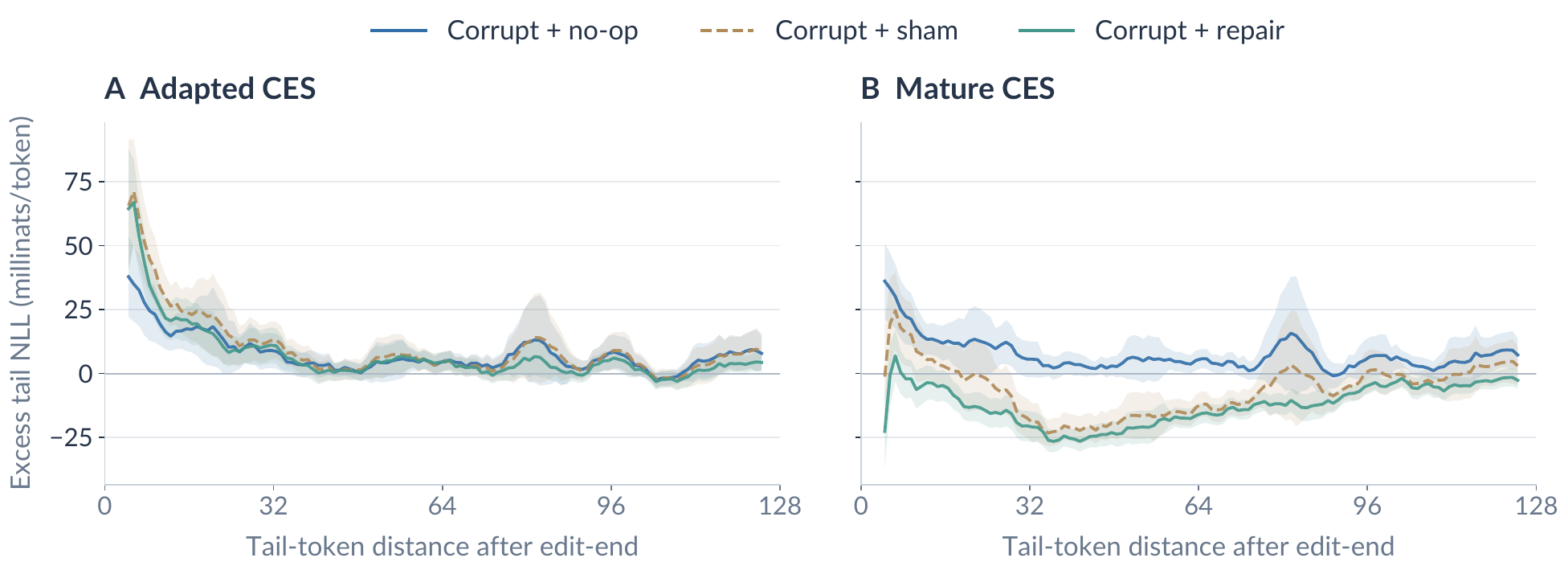}
\caption{Excess tail NLL by distance after edit-end, shown as an eight-token
moving mean with pointwise 95\% uncertainty bands. The mature checkpoint's
gold-repair curve remains below the corrupt empty-program curve throughout
most of the continuation.}
\label{fig:postedittailposition}
\end{figure}

Mature CES improves on the corrupt empty-program condition by
16.58--25.37 millinats/token across the nine operation, span-length, and
distance aggregates, with all 95\% intervals above zero
(Figure~\ref{fig:postedittailoverview}).
Figure~\ref{fig:postedittailposition} shows how the likelihood differences vary
across the continuation.

After 1,000 adaptation steps, clean plain-prefix BPB remains close to the
initial AR checkpoint (0.8173 versus 0.8163), but the supplied repair does not
improve aggregate NLL over the corrupt empty-program condition
(Table~\ref{tab:posteditnatural}). Longer adaptation is not tested.

Mature CES achieves recovery above 100\% because repair-conditioned NLL is
lower than clean-prefix NLL with an empty program. This ratio includes the
effect of appending a non-empty program; the correct repair also lowers NLL
by 6.41 millinats/token relative to the matched sham. The evaluation
uses synthetic corruptions and one checkpoint each for adapted and mature
CES, without testing training-seed variation or model-generated corruptions.

In a supplementary 60-item CORE-derived pilot, repair raises candidate accuracy
from 0.350 to 0.383 and mean clean-answer margin by 0.1267 nats (95\% paired
bootstrap CI $[0.0278,0.2288]$), with two helpful decision flips and none harmful
(OpenBookQA: 0.25 to 0.35 on 20 items), although low clean accuracy, synthetic
corruptions, and exact greedy matching limit interpretation as a CORE result.

\FloatBarrier
\section{Representation, Editing, and Continuation Costs}
\label{sec:systemscomparison}

CES can express sparse revisions without regenerating unchanged text.
The cost depends on how much is changed and whether subsequent generation
retains the edit program or uses materialized edited text. Representation
costs, measured editing latency, and cache rebuilding are considered separately.

\subsection{Representation and Retained-Context Costs}

For a source of $N$ tokens, regeneration appends a complete target of $M$
tokens, whereas CES appends a program of $E$ tokens. Retaining the history
therefore requires approximately $N+M$ versus $N+E$ tokens, plus prompt
controls. Sparse edits can make $E$ much smaller than $M$; dense rewrites
need not provide this advantage.

Figure~\ref{fig:methodscale}G uses the same 485 CoEdIT pairs for three accounts:
\begin{itemize}
\item \textbf{Reference encoding:} tokens required to express the gold revision
in each native format, including delimiters. CES and the DSLs use a common
deterministic token-Levenshtein alignment; regeneration emits the target,
and FIM emits its gold changed envelope.
\item \textbf{Observed output:} tokens generated by d24 seed-42 models after
supervised CoEdIT fine-tuning, including failed attempts. Short outputs can
reflect immediate termination or failed editing.
\item \textbf{Prompt plus minimum program:} the shortest reference completion
under fixed BPE boundaries and prompt templates, plus its prompt. The search
proceeds left to right through source and target, allowing kept tokens,
insertions, deletions and replacements, but no copy or move operations. It
minimizes encoded program tokens, including operation and address overhead,
rather than the number of token edits.
\end{itemize}

Prefilled opening controls count towards the prompt in observed decoding, but
towards the full program in the \emph{reference encoding} series. FIM reference costs
assume correct localization; predicted FIM also includes locator context.
Iterative Levenshtein bars measure sequence storage, not autoregressive output
or refinement work. Reference encodings recover the same target; observed
outputs differ in quality and coverage. Neither account measures latency.

Split dual-index RoPE leaves transformer KV width unchanged. At depth 24 in
bfloat16, transformer keys and values require 144 KiB per retained token;
the auxiliary address keys add 3 KiB per token. This key-only accounting
excludes weights, working activations and unused cache capacity. Shorter
histories can therefore offset the auxiliary storage.

\subsection{Measured Editing Latency}

On a controlled-edit timing set (250 paired inputs per training seed, three
seeds, H100), depth-24 median generation time is 308 ms for CES versus 438 ms
for regeneration. Mean times are 339 and 1,897 ms: the larger gap is strongly
affected by failed regeneration attempts reaching the 640-token cap, compared
with 128 for CES. Correction quality differs between systems, so these are not quality-matched
speedups. Rebuilding is measured separately below.

\FloatBarrier
\subsection{Cache Rebuilding and Resumed Generation}

Returning the edited text or continuing from the source-plus-program history
requires no cache rebuild. If generation instead resumes with the edited text
replacing that history, cached states from the earliest change onward must be
recomputed; even a small early edit can invalidate a long suffix
(Appendix~\ref{sec:inference_details}). We benchmark rebuilding the entire
edited prefix, without reusing its unchanged portion.
Table~\ref{tab:d24cachescaling} compares this rebuild followed by generation
against continuation from an already-cached source.

\begin{table}[htbp]
\centering
\small
\begin{tabular}{rrrrr}
\toprule
Context & Rebuild & Cached continuation & \multicolumn{2}{c}{Extra time (\%)} \\
(tokens) & (ms) & 32 tokens (ms) & 32 tokens & 256 tokens \\
\midrule
128 & 5.9 & 172.4 & 3.4 & 0.6 \\
512 & 13.9 & 172.6 & 8.0 & 1.1 \\
1,792 & 33.0 & 171.6 & 19.5 & 2.3 \\
\bottomrule
\end{tabular}

\caption{\textbf{Full cache rebuilding before continuation, d24 on RTX 5080.}
Cached continuation measures generation from an already-prefilled source,
excluding prefill and rebuilding. Rebuild measures full prefill of the edited
context. Times are medians over three case medians at each length; the first
two timing columns use the 32-token continuation condition. Extra time is the mean paired percentage
increase of rebuild-plus-edited-text continuation over cached source
continuation, computed from per-case median times.}
\label{tab:d24cachescaling}
\end{table}

Nine length-preserving replacement cases cross context lengths of 128, 512,
and 1,792 tokens with three fixed span/location pairs: 1/0.1, 16/0.5, and
64/0.9 (span in tokens; location as a fraction of the allowable start-position
range). Locations are not sampled or varied independently of span size.
Continuation lengths are 32 and 256 tokens.

The pretrained d24 checkpoint (model/data seeds 42/1234) runs at batch size
one, bfloat16, PyTorch 2.9.1, and cache capacity 2,048. Eager inference retains
compute-dtype weights, avoids GPU-to-host cache-position reads, projects only
the final prompt state through the vocabulary head, and scores addresses only
when needed. Each case/condition uses one warmup and three synchronized
repetitions. Timings exclude loading, allocation, tokenization, program
generation and application; rebuild time covers edited-prefix prefill.

Rebuild time increases with context length, while its relative cost falls
with longer continuations. Multiple operations can share one rebuild;
overhead depends on materialization frequency. Repeated edit--resume cycles
and longer contexts remain untested, so these measurements do not establish
an end-to-end editing speedup.

\FloatBarrier
\section{Post-Training Recipe and Evaluation}\label{sec:rlrecipe}

A pretrained depth-24 \CES{} model is post-trained for grammatical error
correction using the three stages in Table~\ref{tab:rlcurriculum}, without
changing its architecture or edit language.

\begin{table}[tbp]
\centering\small
\begin{tabularx}{\linewidth}{l r X}
\toprule
Stage & Updates & Data mixture \\
\midrule
Balanced SFT & 800 & 40\% human edit programs, 40\% matched clean no-ops, 10\% synthetic repairs, 10\% clean continuations. \\
Broad RL & 100 & 50\% model-generated corruptions, 30\% random corruptions, 20\% clean no-ops. \\
GEC RL & 200 & Localized GPT-5 Mini corruptions of clean \CoEdit{} training targets. \\
\bottomrule
\end{tabularx}
\caption{Post-training curriculum. SFT uses 40 examples per global update; each RL update uses 32 prompts with eight sampled programs per prompt. RL data mixtures describe rollout prompts; auxiliary supervised losses are specified below.}
\label{tab:rlcurriculum}
\end{table}

\paragraph{Supervision and pointer anchor.}
Supervised edit loss covers only the completion after $\mathcal{E}_{\text{S}}$,
with address supervision at pointer decisions. SFT exposes 12,800 human
examples, approximately one pass over the 12,756-example training pool.
Editing prompts take the form $(\mathcal{E}_{\text{B}},x,\mathcal{E}_{\text{S}})$,
with the boundary marker at serialized position zero. It anchors insertions
before the first source token but is excluded from deletion/replacement spans.
The marker and source are conditioning inputs.

\paragraph{RL data and rollout settings.}
Broad RL draws without replacement from 359,302 eligible pretraining sources.
GEC RL uses the filtered corruption corpus described in
Appendix~\ref{sec:dataquality}.
Neither the \CoEdit{} validation nor test set contributes training examples.
Both stages use learning-rate scale 0.005; token/pointer temperatures are
0.5/0.5 for broad RL and 0.7/0.7 for GEC RL. Rollouts allow 64 completion
tokens, 32 payload tokens, and two operations. GEC RL applies a closure bias
of 0.25 after at least one payload token, consistently in rollout and replay.

\paragraph{Frozen evaluation partitions.}\label{sec:rlsplits}
The 19,823 GEC rows in the upstream \CoEdit{} training file yield 19,756 unique
normalized source--target pairs. Sorting by deterministic hash (seed
\texttt{ces-rho-gec-pareto-v1}) assigns the first 2,000 to model selection,
the next 5,000 to confirmation, and the remaining 12,756 to human
post-training. The official 485-example validation split is used separately
for auxiliary evaluations. Confirmation decoding and acceptance receive only
the source, never the reference edit or target.

Partitions are pair-disjoint but not source-disjoint: 23 confirmation rows
share a source with a human-training row having a different target, and one
with the selection set. Section~\ref{sec:rljudgingsensitivity} reports their
exclusion. The 5,000 matched identity inputs use confirmation targets as both
source and target, $(y^*,y^*)$; changes measure editing of nominally clean
text, not independently adjudicated semantic errors.

\paragraph{Verifiable reward and policy objective.}
For a corrupted sequence $x$, known clean target $y^*$, and sampled program
$a$, the \CES{} interpreter produces $\hat y$.  Let
\begin{equation}
 \Delta_k=\operatorname{clip}\!\left(
 \frac{d_k(x,y^*)-d_k(\hat y,y^*)}{\max(d_k(x,y^*),1)},-1,1\right),
 \qquad k\in\{\mathrm{tok},\mathrm{char}\},
\end{equation}
where $d_k$ is token- or character-level Levenshtein distance.  For a valid
program define
\begin{equation}
 s(a)=0.9\Delta_{\mathrm{tok}}+0.1\Delta_{\mathrm{char}}
 +0.25\,\mathbf{1}[\hat y=y^*]
 -0.1\frac{d_{\mathrm{tok}}(x,\hat y)}{\max(|x|,1)}
 -0.03\,\mathbf{1}[x\ne y^*\wedge\hat y=x].
\end{equation}
The terminal reward is $-1$ for an invalid program, $1.25s(a)$ when a valid
program has $s(a)<0$, and $s(a)$ otherwise. Reward uses executable recovery,
not an online judge. Broad RL includes clean cases with directly verifiable
no-op execution; GEC RL instead teaches clean behavior through supervised
no-op and first-decision examples.

For each prompt's $G=8$ sampled programs, the unstandardized leave-one-out advantage is
\begin{equation}
 A_i=r_i-\frac{1}{G-1}\sum_{j\ne i}r_j .
\end{equation}
The policy loss replays each sampled trajectory and sums the log probabilities
of its genuinely sampled token and pointer decisions $\mathcal D_i$:
\begin{equation}
 \mathcal L_{\mathrm{RL}}
 =-\frac{1}{\sum_i|\mathcal D_i|}
 \sum_i A_i\sum_{t\in\mathcal D_i}\log\pi_\theta(a_{i,t}\mid h_{i,t}).
\end{equation}
Forced grammar actions do not enter the loss, and both decoding and replay use
the same legal-token and valid-pointer masks.  This is an on-policy leave-one-out REINFORCE objective \citep{ahmadian2024rloo} with decision normalization and without PPO clipping.

Broad RL mixes 40\% on-policy loss, 10\% synthetic completion loss,
10\% clean autoregressive loss, 20\% human completion loss, and 20\% clean
no-op loss. GEC RL uses 45\% on-policy, 20\% synthetic completion,
20\% human completion, 10\% clean autoregressive, and 5\% paired
first-decision loss. The latter groups Delete, Insert, and Replace openers
into an ``edit'' class contrasted with immediate $\mathcal{E}_{\text{E}}$, with
training edit prior 0.35.

\paragraph{Target-blind inference gate.}
At the first program decision, an edit is allowed when the log-sum-exp of
Delete, Insert, and Replace opener logits minus the $\mathcal{E}_{\text{E}}$ logit is at least $\tau$; otherwise the
input is returned. Equivalently, the combined opener probability must be at
least $e^\tau$ times the end probability, rather than any individual opener
being most probable. The filtered policy uses $\tau=0.5$, with $\tau=1.0$
also reported. The gate uses neither targets nor clean-model likelihood.

\paragraph{Full confirmation acceptance policy.}
Figure~\ref{fig:rlpolicy}D--E additionally requires at most four
removed-plus-inserted characters, one changed region, and program confidence
of at least $-0.6$. A character-level \texttt{SequenceMatcher} comparison
identifies changed regions; replacing two characters with two others counts
as four. Confidence averages token and pointer log probabilities, excluding
grammar-forced decisions, from replay under legal masks at temperature 1
without top-$k$ or nucleus truncation. Any failed check returns the source.
Greedy evaluation allows 96 completion tokens, 32 payload tokens, and four
operations. All 5,000 confirmation candidates parse successfully; filtering
leaves 880 changed outputs and 74 changes on the matched identity inputs.

\subsection{Confirmation judging and sensitivity}\label{sec:rljudgingsensitivity}
\texttt{gpt-5-mini-2025-08-07} judges anonymous, shuffled model and human
candidates, without treating the human edit as authoritative. ``Improves''
means a net correction preserving intent; ``same'' covers cosmetic or
net-neutral changes; ``worse'' means material linguistic or semantic harm. Unchanged
outputs receive ``same''; identical candidates cannot receive sole-method
preference. Intervals use 20,000 example-bootstrap resamples (seed 42).

Excluding the 24 repeated-source rows changes improvement/harm/net improvement
from 12.02/2.76/$+9.26$ to 12.06/2.73/$+9.32$ percent on 4,976 cases.
Results concern one selected d24 checkpoint and do not isolate an RL
advantage: there is no full-curriculum, acceptance-filter-matched SFT
comparison or judged regeneration baseline on this set. Bootstrap intervals
exclude checkpoint selection, judge-family uncertainty and evaluation-set
reuse; GPT-5 Mini both generates training corruptions and judges outputs.
Auxiliary validation did not show consistently positive net judged quality
across depths and seeds, and judgments varied with accompanying candidates;
robustness across checkpoints and judging comparisons remains unestablished.

\subsection{Unfiltered decoding and threshold sweep}\label{sec:rlunfiltered}
The same 5,000 sources and 5,000 identity inputs use unchanged grammar masks
and decoding limits, without acceptance filters. Plain greedy uses zero
token/pointer temperatures and no entry gate. The threshold sweep generates
one greedy candidate with a forced initial edit, then uses the source-only
entry margin to select it or the unchanged input. This agrees with direct
decoding in all 128 checks (four thresholds, 16 inputs per split).
The sweep covers $\tau=-5$ to $3$ and forced editing; plots show $\tau\ge-1$,
below which coverage saturates. Invalid programs return the source and remain
in all-input denominators.

The 2,983 unique changed input/output pairs use the judging protocol above,
sharing labels for identical outputs on the same input. Unchanged outputs
are neutral; identity-input changes use exact text comparison. Filtered and
unfiltered outputs are judged separately, so Figure~\ref{fig:rlpolicy}'s rows
are not paired estimates of filtering effects.

\FloatBarrier
\section{Shared Pointer Tokens and Further Directions}
\label{sec:sharedpointer}
A matched single-seed d12 ablation replaces all 11 role-specific pointers
with the insertion-location token, retaining grammar-dependent roles and
masks, vocabulary size, and parameter count. Each arm receives 26,880 updates,
854,219,434 raw tokens, 1,342,019 examples, and 5,355,059 pointer targets,
taking approximately 115 minutes on two RTX 5080s.
\begin{table}[tbp]
\centering\small
\begin{tabular}{lrr}
\toprule
Metric & Role-specific & Shared token \\
\midrule
Validation BPB & 0.905563 & 0.904688 \\
Synthetic pointer accuracy & 48.64\% & 45.81\% \\
Controlled program validity & 87.2\% & 88.0\% \\
Controlled exact recovery ($n=250$) & 2.4\% (6) & 3.6\% (9) \\
CoEdIT pointer accuracy & 23.12\% & 23.34\% \\
\bottomrule
\end{tabular}
\caption{Shared-pointer ablation. Three extra exact repairs do not establish a robust advantage.}
\label{tab:sharedpointer}
\end{table}
Table~\ref{tab:sharedpointer} shows that sharing preserves similar language-modeling and editing scores, while
role-specific tokens improve synthetic localization by 2.83 percentage points.
Transfer to new operations or instruction-defined pointer roles remains untested.

Separate parameter groups or routed experts could specialize in address
prediction. Learned corruption policies could generate hard but recoverable
edits, constrained by realism, ambiguity, and edit budgets to preserve intent
and prevent reward exploitation. Draft--revision training could target useful
improvements beyond corruption inversion, compared with compute-matched
regeneration and self-refinement. These directions remain untested.

CES could update an agent's working context, but compaction requires
materializing revised text and rebuilding its cache, or external storage;
appending programs alone retains old tokens.

\clearpage
\newpage
\section*{NeurIPS Paper Checklist}
\begin{enumerate}

\item {\bf Claims}
    \item[] Question: Do the main claims made in the abstract and introduction accurately reflect the paper's contributions and scope?
    \item[] Answer: \answerYes{}
    \item[] Justification: The abstract and introduction distinguish the CES representation, completed baseline evidence, and selective post-training extension; limitations are explicit in Section~\ref{sec:limitations}.

\item {\bf Limitations}
    \item[] Question: Does the paper discuss the limitations of the work performed by the authors?
    \item[] Answer: \answerYes{}
    \item[] Justification: Section~\ref{sec:limitations} discusses model scale, baseline strength, training-data coverage, selective editing, judge uncertainty, repeated sources, and cache-rebuilding costs.

\item {\bf Theory assumptions and proofs}
    \item[] Question: For each theoretical result, does the paper provide the full set of assumptions and a complete (and correct) proof?
    \item[] Answer: \answerNA{}
    \item[] Justification: The paper does not present formal theorems or proofs. It is primarily an architectural and empirical study.

    \item {\bf Experimental result reproducibility}
    \item[] Question: Does the paper fully disclose all the information needed to reproduce the main experimental results of the paper to the extent that it affects the main claims and/or conclusions of the paper (regardless of whether the code and data are provided or not)?
    \item[] Answer: \answerYes{}
    \item[] Justification: The architecture and protocols are specified in the paper, with public code, checkpoints, configurations, and data reconstruction instructions linked in ``Software and Data''. The repository documentation states the remaining reproduction limits.

\item {\bf Open access to data and code}
    \item[] Question: Does the paper provide open access to the data and code, with sufficient instructions to faithfully reproduce the main experimental results, as described in supplemental material?
    \item[] Answer: \answerYes{}
    \item[] Justification: Public code and artifacts are linked in ``Software and Data'', with installation, data preparation, training, and evaluation instructions. CoEdIT inputs are reconstructed locally from upstream data; confirmation predictions and judgments, including explanations, are released with reconstruction and viewing tools.

\item {\bf Experimental setting/details}
    \item[] Question: Does the paper specify all the training and test details (e.g., data splits, hyperparameters, how they were chosen, type of optimizer) necessary to understand the results?
    \item[] Answer: \answerYes{}
    \item[] Justification: Appendices~\ref{sec:reproducibility} and~\ref{sec:rlrecipe} specify architectures, optimizers, learning-rate schedules, batch and update budgets, training seeds, split construction, evaluation policies, and the post-training curriculum. Public configurations and lineage documentation accompany the code and checkpoints.

\item {\bf Experiment statistical significance}
    \item[] Question: Does the paper report error bars suitably and correctly defined or other appropriate information about the statistical significance of the experiments?
    \item[] Answer: \answerYes{}
    \item[] Justification: Main method tables report sample standard deviations over three paired training seeds. Continuation and semantic estimates give example-bootstrap intervals; the semantic confirmation results concern one selected post-trained checkpoint and do not establish robustness across training seeds.

\item {\bf Experiments compute resources}
    \item[] Question: For each experiment, does the paper provide sufficient information on the computer resources (type of compute workers, memory, time of execution) needed to reproduce the experiments?
    \item[] Answer: \answerYes{}
    \item[] Justification: Appendix~\ref{sec:reproducibility} gives GPU types and memory, approximate training and post-training runtimes, recorded training GPU-hours, and API costs. These estimates describe reproduction requirements; additional preliminary experiments are acknowledged rather than included in the recorded totals.
    
\item {\bf Code of ethics}
    \item[] Question: Does the research conducted in the paper conform, in every respect, with the NeurIPS Code of Ethics \url{https://neurips.cc/public/EthicsGuidelines}?
    \item[] Answer: \answerYes{}
    \item[] Justification: To the best of our knowledge, the work conforms to the NeurIPS Code of Ethics. The work uses existing datasets and generated edits for research purposes and includes an impact discussion; see the ``Impact Statement'' section.

\item {\bf Broader impacts}
    \item[] Question: Does the paper discuss both potential positive societal impacts and negative societal impacts of the work performed?
    \item[] Answer: \answerYes{}
    \item[] Justification: The paper includes a dedicated impact statement discussing positive uses in controllable editing and iterative refinement, as well as misuse risks related to subtle document manipulation and code rewriting; see the ``Impact Statement'' section.
    
\item {\bf Safeguards}
    \item[] Question: Does the paper describe safeguards that have been put in place for responsible release of data or models that have a high risk for misuse (e.g., pre-trained language models, image generators, or scraped datasets)?
    \item[] Answer: \answerNA{}
    \item[] Justification: The released models are small research prototypes with limited editing capabilities, and the release is not expected to pose a high risk of misuse. The public data exclude credential-screened rows and retain applicable upstream restrictions.

\item {\bf Licenses for existing assets}
    \item[] Question: Are the creators or original owners of assets (e.g., code, data, models), used in the paper, properly credited and are the license and terms of use explicitly mentioned and properly respected?
    \item[] Answer: \answerYes{}
    \item[] Justification: Upstream creators are cited, and component licences and attribution accompany the public repositories. NanoChat MIT attribution and ClimbMix source-text terms are retained; CoEdIT inputs are reconstructed locally rather than redistributed.

\item {\bf New assets}
    \item[] Question: Are new assets introduced in the paper well documented and is the documentation provided alongside the assets?
    \item[] Answer: \answerYes{}
    \item[] Justification: The public repositories document the data, tokenizer, and checkpoints, including component licences, checksums, model lineages, usage instructions, and release limitations.

\item {\bf Crowdsourcing and research with human subjects}
    \item[] Question: For crowdsourcing experiments and research with human subjects, does the paper include the full text of instructions given to participants and screenshots, if applicable, as well as details about compensation (if any)? 
    \item[] Answer: \answerNA{}
    \item[] Justification: The paper does not report crowdsourcing experiments or new research with human subjects. It evaluates on existing text-edit datasets.

\item {\bf Institutional review board (IRB) approvals or equivalent for research with human subjects}
    \item[] Question: Does the paper describe potential risks incurred by study participants, whether such risks were disclosed to the subjects, and whether Institutional Review Board (IRB) approvals (or an equivalent approval/review based on the requirements of your country or institution) were obtained?
    \item[] Answer: \answerNA{}
    \item[] Justification: The paper does not involve new research with human subjects or crowdsourcing conducted by the authors.

\item {\bf Declaration of LLM usage}
    \item[] Question: Does the paper describe the usage of LLMs if it is an important, original, or non-standard component of the core methods in this research? Note that if the LLM is used only for writing, editing, or formatting purposes and does \emph{not} impact the core methodology, scientific rigor, or originality of the research, declaration is not required.
    \item[] Answer: \answerYes{}
    \item[] Justification: The paper describes GPT-5 Mini generation of edit/corruption data and reference-blind semantic judging, including the shared-family limitation. No online LLM judge is used as the selected RL reward.

\end{enumerate}

\end{document}